\documentclass[10pt,journal,compsoc]{IEEEtran}
\def\degree{${}^{\circ}$}
\ifCLASSOPTIONcompsoc
  \usepackage[nocompress]{cite}
\else
  \usepackage{cite}
\fi

\ifCLASSINFOpdf

\else

\fi

\usepackage{float}
\usepackage{amsmath}
\usepackage{graphicx}
\usepackage{caption}
\usepackage{stfloats}
\usepackage{indentfirst} 
\usepackage{multirow}
\usepackage{booktabs}
\usepackage{bigstrut}
\usepackage{hyperref}
\usepackage[table]{xcolor}
\usepackage{subfigure}
\usepackage{algorithm}
\usepackage{algorithmic}
\usepackage{amssymb}
\usepackage{longtable}
\usepackage{cite}
\usepackage{makecell}
\usepackage{soul}
\usepackage[numbers,sort,compress]{natbib}

\usepackage{amsthm,amsmath,amssymb}
\usepackage{mathrsfs}

\begin{document}

\title{Towards Domain-Independent and Real-Time Gesture Recognition Using mmWave Signal}

 \author{Yadong~Li,
         Dongheng~Zhang,
         Jinbo~Chen,
         Jinwei~Wan,
         Dong~Zhang,
         Yang~Hu,
         Qibin~Sun,~\IEEEmembership{Fellow,~IEEE,}
         and~Yan~Chen,~\IEEEmembership{Senior Member,~IEEE}% <-this % stops a space    
\IEEEcompsocitemizethanks{\IEEEcompsocthanksitem Y. Li, D. Zhang, J. Chen, Q. Sun, Y. Chen are with the School
of Cyber Science and Technology, University of Science and Technology of China, Hefei 230026, China
(E-mail: yadongli@mail.ustc.edu.cn, dongheng@ustc.edu.cn, jinbochen@mail.ustc.edu.cn, qibinsun@ustc.edu.cn, eecyan@ustc.edu.cn).
\IEEEcompsocthanksitem J. Wan is with the China Nanhu Academy of Electronics and Information Technology, Jiaxing 314002, China
(E-mail: wan3137@163.com).
\IEEEcompsocthanksitem D. Zhang is with the Institute of Advanced Technology, University of Science and Technology of China, Hefei 230031, China (E-mail: zdtop@iat.ustc.edu.cn).
\IEEEcompsocthanksitem Y. Hu is with the School of Information Science and Technology, University of Science and Technology of China, Hefei 230026, China (E-mail: eeyhu@ustc.edu.cn).}% 
}

%{Shell \MakeLowercase{\textit{et al.}}: Bare Demo of IEEEtran.cls for Computer Society Journals}

\IEEEtitleabstractindextext{%
\begin{abstract}

Human gesture recognition using millimeter-wave (mmWave) signals provides attractive applications including smart home and in-car interfaces. While existing works achieve promising performance under controlled settings, practical applications are still limited due to the need of intensive data collection, extra training efforts when adapting to new domains, and poor performance for real-time recognition. In this paper, we propose DI-Gesture, a domain-independent and real-time mmWave gesture recognition system.
Specifically, we first derive signal variations corresponding to human gestures with spatial-temporal processing. To enhance the robustness of the system and reduce data collecting efforts, we design a data augmentation framework for mmWave signals based on  correlations between signal patterns and gesture variations. Furthermore, a spatial-temporal gesture segmentation algorithm is employed for real-time recognition. Extensive experimental results show DI-Gesture achieves an average accuracy of 97.92\%, 99.18\%, and 98.76\% for new users, environments, and locations, respectively. We also evaluate DI-Gesture in challenging scenarios like real-time recogntion and sensing at extreme angles, all of which demonstrates the superior robustness and effectiveness of our system.

\end{abstract} 

}

\maketitle
\IEEEdisplaynontitleabstractindextext

\IEEEpeerreviewmaketitle

\IEEEraisesectionheading{\section{Introduction}\label{sec:introduction}}
\IEEEPARstart{H}{uman} gesture recognition plays an important role in human-computer interface systems, which provides users a more natural and convenient way to interact and control machines and devices. 
For instance, in smart homes\cite{iot}, people can control household Internet of Things (IoT) devices with gestures in a contactless way, which provides entertaining user experience. 

Traditional approaches for gesture recognition are based on cameras \cite{camera1,camera2} or wearable sensors\cite{smartwatch,smartwatches}. 
Although these techniques have achieved impressive recognition accuracy, their deployments in real-world applications still remain challenges. Camera-based solutions have to deal with illumination variations \cite{light} and privacy issues \cite{privacy} while wearable sensors require physical contact between the human body and device \cite{wearable}, which is uncomfortable and not suitable for long-term use. To resolve these challenges, the recent wireless sensing technique has demonstrated its ability for contactless sensing, including vital sign monitoring \cite{vital1,vital3,speednet,mtrack}, gait recognition \cite{gait1,gait2,gait3}, human identification  \cite{HumanRecognition}, pose estimation \cite{pose1,pose2,pose3,pose4} and human activity recognition \cite{wifall,radhar,wihar}. Compared with traditional sensing methods, the wireless sensing technique is more privacy-friendly and robust under different illumination conditions. 
In the past years, gesture recognition based on different wireless mediums, including WiFi \cite{widar,wifigesture1,wifivision,eventsdetection,phaseoffset1,phaseoffset2,wigest}, acoustic signals \cite{acoustic} and millimeter-wave \cite{latern} has been investigated. 
Among these mediums, millimeter-wave draws lots of attention due to its significant advantages. First of all, the fine-grained range and angle resolution of mmWave radar enable sensing of subtle motion. Secondly, the anti-interference ability of mmWave signal is strong due to its high frequency band.
Finally, the small size of mmWave radar chip \cite{soli} make it easy for being embedded in portable devices. 
Hence, many research efforts have been made to exploit mmWave signals for gesture recognition \cite{tinyradarnn,soli,radarnet,pointcloud}. For instance, deep-soli \cite{soli} achieves fine-grained gesture recognition with a compact mmWave radar and deep neural network. RadarNet \cite{radarnet} designs an efficient neural network and collects a large-scale dataset including over 4 million samples to train a robust model.

However, existing methods are limited in three aspects. ($i$) Labor-intensive data collection. To ensure the robustness of deep learning models, researchers have to collect sufficient training data to prevent overfitting, which is tedious and impractical. 
Note that there are few large-scale mmWave gesture datasets public to the research community, making data collection a more annoying problem.  
($ii$) Domain dependence. Deep learning-based approaches achieve high accuracy when training and testing models under familiar domains. However, model retraining or extra-training efforts are still required when adapting to new domains since the propagation of mmWave signals is subject to change upon the variation of environment setup, relative locations, and gesture speed of individuals.   ($iii$) Off-line recognition and poor performance in real-time scenarios. Most existing works focus on off-line gesture recognition, assuming that all gesture samples are well-segmented before passing into the classifier. However, in practical scenarios, the system operates in real-time, which is more difficult than the segmented classification task.

\begin{figure*}[t]
\begin{center}
\includegraphics[scale=0.18]{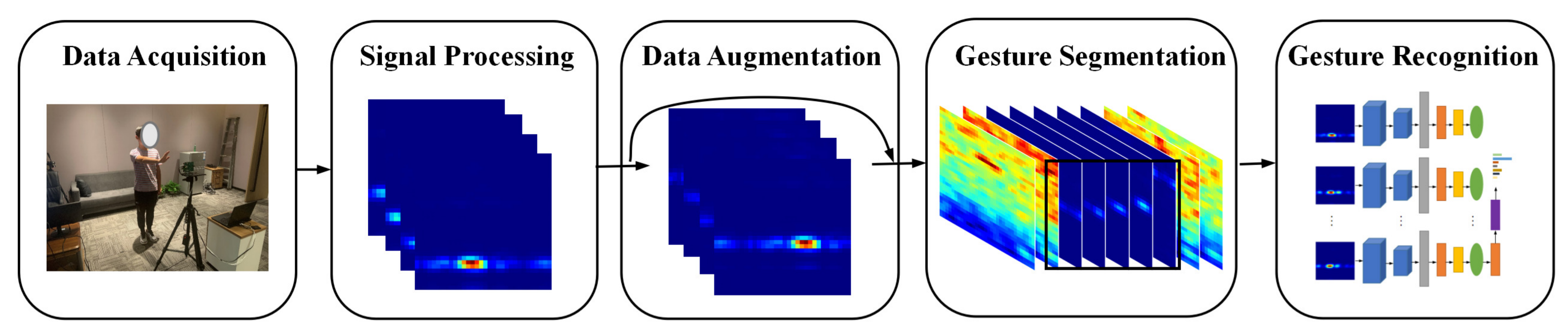}
\caption{Overview structure of DI-Gesture. DI-Gesutre transforms received signals into DRAI sequences and generates synthetic data by the proposed data augmentation framework. For real-time recognition, DI-Gesutre detects gesture boundaries from the continuous DRAI stream. Finally, the segmented samples are fed into neural networks for classification.}   
\label{overview}
\end{center}
\end{figure*}

One of the most efficient techniques to make models generalize better with insufficient training data is data augmentation \cite{deeplearning}. While lots of research efforts have been devoted to designing effective data augmentation methods tailored to different tasks \cite{augmentation1,augmentation2,augmentation3}, existing methods could not be directly applicable to mmWave signals for two reasons:
$(i)$ unlike optical images, the semantic information of mmWave signals is not interpretable intuitively; $(ii)$ the signal characteristic (i.e. angular resolution, radiated power) vary with different directions, which are major peculiarities of mmWave radar leading to severe performance degradation when sensing at extreme angles.

In this article, we propose DI-Gesture, a real-time mmWave gesture recognition system which can generalize to gestures performed by new users, at new locations or in new environments with high accuracy and low latency. As shown in Fig. \ref{overview}, to extract robust features, we first derive spatial-temporal variations of gestures while reducing environmental discrepancy by transforming raw radar signals to Dynamic Range Angle Image (DRAI). Then, by analyzing signal characteristics and observing correlations between signal patterns and gesture variations in different situations, we note that most of gesture variations can be simulated by applying certain transformations to the original data. Therefore, we propose a data augmentation framework considering characteristics of mmWave signals to increase the diversity of the dataset, reduce data collecting efforts, and enhance the robustness of neural networks across various domains.

To make the system work in real-time and in the presence of interfering users, we design a spatial-temporal segmentation algorithm based on CFAR \cite{CFAR} and CLEAN \cite{clean} to separate signals of multiple users in the spatial domain, then detect gesture boundaries of DRAI sequences in the temporal domain. Finally, we design a lightweight deep neural network model to extract frame-level and sequence-level features from DRAI sequences and perform gesture classification.

To evaluate DI-Gesture, we implement it on a commodity radar sensor and collect 24050 samples of 12 common gestures from 25 participants, 5 locations, and 6 environments. Experimental results show that DI-Gesture achieves an average accuracy of 97.92\%, 99.18\%, and 98.76\% for new users, environments, and locations, respectively. When sensing at extreme angles, the proposed data augmentation framework can bring up to 22.50\% of accuracy improvement. 
For real-time recognition, the average accuracy and inference time of DI-Gesture is 97.08\% and 2.87ms, respectively. Moreover, we compare DI-Gesture with state-of-the-art solution, and results show that our system significantly outperforms the state-of-the-art under cross-domain conditions and real-time scenarios.

We summarize the main contributions of this article as follows.

(1) To the best of our knowledge, we are the first to address the domain dependence problem of mmWave radar gesture recognition in various domains (i.e. environments, users, distances, and angles).

(2) We design a data augmentation framework for mmWave signals based on correlations between signal representations and gesture variations to ease the pain of data collection, improve the robustness of the classifier and enhance the ability of sensing at extreme angles.

(3) We implement DI-Gesture on a commodity mmWave radar and conduct extensive evaluations. Experiment results demonstrate the impressive performance of DI-Gesture under cross-domain settings and real-time scenarios.

(4) We collect and label the first comprehensive cross-domain mmWave gesture dataset\footnote[1]{https://github.com/DI-HGR/cross\_domain\_gesture\_dataset}, consisting of 24050 samples from 25 volunteers, 5 locations and 6 environments, which has been public to the research community. We believe that this dataset would facilitate future research of mmWave gesture recognition.

\section{Related Work}
Wireless sensing is a promising and interesting technique which has various IoT applications and lots of research has been conducted to exploit different types of wireless signals including mmWave signals \cite{latern,tinyradarnn,soli,radarnet,pointcloud} and WiFi signals \cite{widar,wifigesture1,wifinger} to recognize human hand gestures. These approaches vary in terms of signal characteristics, data preprocessing and neural network structure. For example, WiFinger \cite{wifinger} achieves fine-grained gesture recognition utilizing the channel state information (CSI) of WiFi, which is however too susceptible to environmental changes. To achieve cross-domain recognition without collecting extra training data, Widar3.0 \cite{widar} extracts domain-independent features of gestures from WiFi signals, thus addressing the domain dependence problem in the feature level.
Beyond WiFi, mmWave gesture recognition has attracted great attention due to the fine-grained resolution, small chip size, and easy to deployment of mmWave sensors. For example, deep-soli \cite{soli} designs a deep neural network containing both CNN and LSTM to model the dynamic information in mmWave signals and achieves fine-grained gesture recognition with a short-range 60GHz radar. Liu. et al. \cite{smarthome} extract dynamic variation of gestures from mmWave signals and design a lightweight CNN to recognize gestures in long-range scenario. Beyond recognizing simple gestures, mmASL \cite{mmasl} extracts frequency features from 60GHz mmWave signals and achieves American sign language recognition with a multi-task deep neural network. To reduce the computational cost, RadarNet \cite{radarnet} proposes an efficient neural network with a model size of only 0.14MB and trains the model on a large-scale dataset to achieve an accuracy of over 99\%.

All aforementioned methods achieve high accuracy on their gesture dataset. However, deep learning-based approaches need a  sufficient amount of training data to enhance the robustness of neural networks. Moreover, original mmWave signals usually contain domain-specific information irrelevant to gestures, which makes it a more challenging problem to adapt a trained classifier to another domain. Several prior works have tried to solve this problem. For instance, RF-Net \cite{rfnet} proposes a meta-learning framework to adapt the trained classifier to new environments with only one labeled sample. Wang et al. \cite{gan} utilize generative adversarial network (GAN) to generate virtual training samples and avoid laborious data collection. 
Another limitation of existing works is that most of them focus on segmented classification tasks that manually label the start and the end of gestures, 
which are not applicable in real-world application scenarios. 
In practice, the gesture recognition system needs to handle continuous signals with unknown gesture boundaries. Liu et al. \cite{smarthome} design a hidden Markov model-based voting mechanism to perform gesture segmentation. RadarNet \cite{radarnet} proposes a sliding window approach for continuous gesture recognition and filters predictions with some experimental heuristics. The fixed-length sliding window approach is straightforward but cannot work well on gestures with diverse speeds, meaning no prediction or multiple predictions for a single gesture.

In this article, we present a domain-independent and real-time gesture recognition system using mmWave signals. Our work is different from prior works in the following three aspects. Firstly, existing works \cite{mgesture,smarthome,pointcloud} are limited to handle some of the domain factors, which are not comprehensive, while we are the first to address the domain dependence problem of mmWave radar gesture recognition across various domains (i.e. environment, user, distances, and angles). Secondly, beyond extracting gesture-relevant information from raw signals like Widar3.0 \cite{widar}, we further design a data augmentation framework for mmWave signals to reduce efforts of data collection and enhance the adaptation ability of the trained model in new domains. Compared with existing works \cite{rfnet,gan}, the data augmentation framework can generate synthetic data without extra training efforts. Finally, in contrast to prior work \cite{tinyradarnn,mmasl,earlygesture} focusing on off-line recognition, we evaluate our system in the real-time scenario which is more challenging. 

\section{System Design}

\subsection{Signal Processing}

\begin{figure}[htbp]
	\begin{center}
		\includegraphics[scale=0.17]{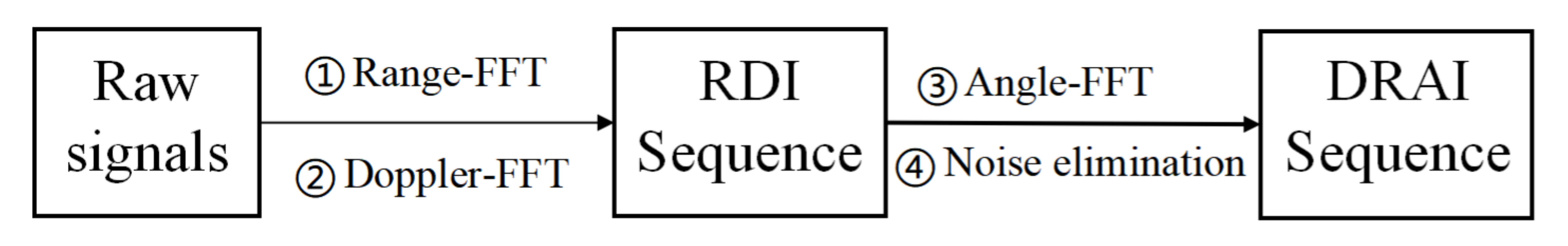}
		\caption{The calculation process of DRAI.}
		\label{process}
	\end{center}
\end{figure}

As shown in Fig. \ref{process}, DI-Gesture employs the Frequency Modulated Continuous Wave (FMCW) radar to obtain Dynamic Range Angle Image (DRAI) for gesture recognition. 
Specifically, we first perform 3D-FFT on raw signals to derive the ranges, velocities, and angles of gestures. Then, we conduct noise elimination to filter environmental interference and improve the robustness of classifiers. 

\emph{(1) Range-FFT:} The radar continuously transmits FMCW signals (i.e. chirps), which will be reflected after hitting the detected object and received by receive antennas. Then the mixer on the radar board will mix the received chirp with the transmitting chirp to obtain an intermediate frequency (IF) signal. The relationship between the frequency of IF signal ${f}$ and the distance $d$ between radar and object can be denoted as 

\begin{equation}
	f = S \cdot \tau = \frac{S \cdot 2d}{c} \Rightarrow d = \frac{fc}{2S} ,
\end{equation}
where $S$ is the slope of the chirp signal, and $c$ is the speed of the signal. Therefore, the range of the detected object can be computed using FFT.

\emph{(2) Doppler-FFT:} To derive the moving speed of the targeted object, the radar transmits a frame that consists of N chirps. The velocity of the object $v$ can be derived from phase differences $\Delta \phi$ caused by doppler effect between two adjacent chirps as
\begin{equation}\label{eqn2}
	\Delta \phi =  \frac{4 \pi v T_c}{\lambda}  \Rightarrow v = \frac{\lambda \Delta \phi}{4 \pi T_c} ,
\end{equation}
where $\lambda$ is the wavelength of the signal, and $T_c$ is the time interval between two adjacent chirps.
According to Eq.~\ref{eqn2}, we perform FFT among N chirps to extract doppler information and obtain Range Doppler Image (RDI).

\emph{(3) Angle-FFT:} The Angle of Arrival (AoA) can be computed by cascading multiple RDIs obtained from different antennas according to phase changes between adjacent receiving antennas \cite{aoa}. The relationship between the phase difference  $\Delta \phi$ and AoA $\theta$ can be derived as 
\begin{equation}
	\Delta \phi =  \frac{2 \pi l \sin \theta}{\lambda},
\end{equation}
where $l$ is the distance between adjacent receiving antennas. 
After Angle-FFT, we have obtained the range-doppler-angle matrix for further processing.

\emph{(4) Noise Elimination:} Since moving targets and static clutters can be discriminated based on doppler frequency, we simply set doppler frequency lower than a velocity threshold as 0 to remove static clutter. To eliminate multipath reflections, we sum the averaged range doppler matrix along the range dimension to obtain the signal intensity of each doppler bin and experimentally set a threshold. Finally, only doppler bins whose signal intensity is higher than the threshold will be counted when summing the range-doppler-angle matrix along doppler dimension into a 2D matrix to obtain DRAI. The detailed description of noise elimination is shown in Algorithm 1.
\begin{algorithm}[H]
	\caption{Noise elimination}
	\begin{algorithmic}[1] 
		\REQUIRE 
		Total number of receiving channels $N$, doppler bin threshold $\tau$, scale factor of the doppler power thershold $\alpha$,
		Range Doppler Images of the receiving channels
		$\{{\rm \bf RD}_i  \in \mathbb{R}^{K\times L} \parallel i=1,\dots,N\}$ 
		\ENSURE  Dynamic Range Angle Image ${\rm {\bf DRAI}}\in \mathbb{R}^{K\times I}$
		\FOR {$ i = 1 $; $ i < N $; $ i ++ $ }
		\STATE {Set ${\rm \bf RD}_i(:,\frac{L}{2}-\tau:\frac{L}{2}+\tau)$ as 0}
		\ENDFOR
		% \FORALL{ ${\rm \bf RD}$}
		%     \STATE{Set ${\rm \bf RD}(:,\frac{L}{2}-\tau:\frac{L}{2}+\tau)$ as 0}
		% \ENDFOR
		\STATE {Get doppler power of each doppler bin ${\rm \bf DP}\in \mathbb{R}^{L}$ by ${\rm \bf DP= Sum (Mean(RD}_1,\dots ,{\rm \bf RD}_i,\dots,{\rm \bf RD}_N))$}
		\STATE {Get the doppler power threshold $T$  by $T = \alpha \times {\rm \bf Max(DP)}$}
		\STATE {Get the range-doppler-angle matrix ${\rm \bf RDA}  \in \mathbb{R}^{K\times L\times I}$ by ${\rm \bf RDA=AngleFFT(RD}_1,\dots ,{\rm \bf RD}_i,\dots,{\rm \bf RD}_N)$}
		
		\STATE {Initialize ${\rm \bf DRAI} \in \mathbb{R}^{K\times I}$ as null matrix and $j=1$}
		
		\WHILE {$ j<L  $}
		\IF {${\rm \bf DP}(j) > T$}
		\STATE {$\rm \bf DRAI$ = $\rm \bf DRAI$ $+$ ${\rm \bf RDA}(:,j,:)$}
		\ENDIF
		\STATE {$j=j+1$}
		\ENDWHILE
		
	\end{algorithmic} 
\end{algorithm}

\begin{figure}[htbp]
	\begin{center}
		\includegraphics[scale=0.18]{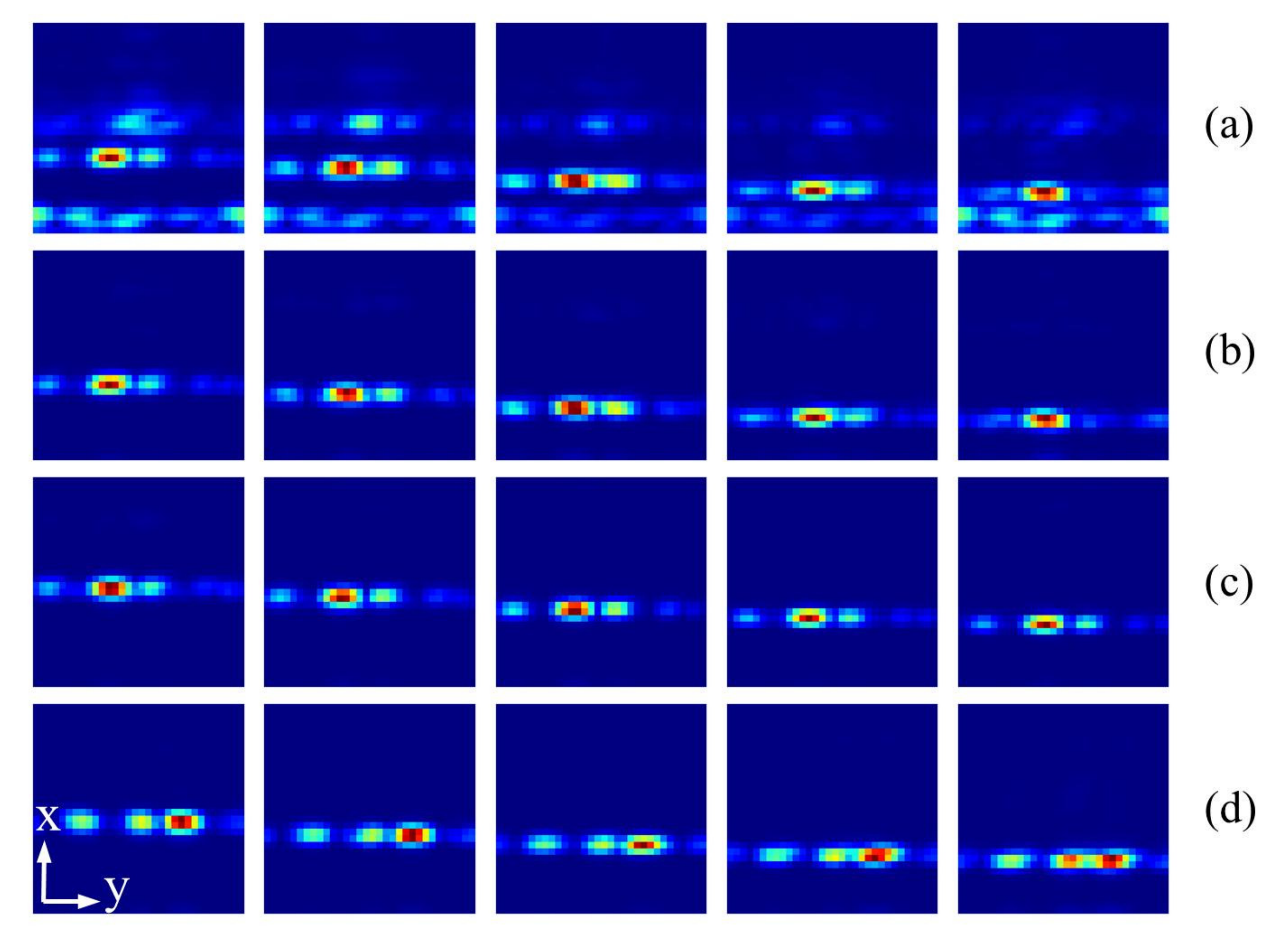}
		\caption{Examples of push at different locations. (a) RAI of push at 60cm; (b) DRAI of push at 60cm; (c) DRAI of push at 80cm; (d) DRAI of push at 30\degree. Columns represent time series of 5 frames. In RAI and DRAI, pixel color, x-axis and y-axis correspond to doppler power,  range, and AoA, respectively.  
		%pixel intensity corresponds to doppler power, horizontal axis is AoA, vertical axis is range.
		}

		\label{push}
	\end{center}
\end{figure}

Fig. \ref{push} shows a series of Range Angle Images (RAI, i.e. directly summing the range-doppler-angle matrix along doppler dimension without noise elimination) and DRAI when users push at different locations. From Fig. \ref{push}(a) and (b) we have two key observations. Firstly, different gestures result in different dynamic patterns in RAI and DRAI. For example, when users perform push, the brightest spot moves vertically which denotes distance changes of hands. Another observation is that compared with RAI, features in DRAI are clearer after static removal and noise elimination.

\subsection{Data Augmentation}
Recently, neural networks have achieved impressive accomplishments in image classification, which motivates us to leverage this technique to build a classifier for mmWave gesture recognition. However, the accuracy and robustness of neural networks are highly dependent on the quantity and quality of training data. Unlike the field of computer vision or natural language processing, the wireless sensing area still lacks large-scale, high-quality, and public mmWave gesture datasets. 

To tackle this problem, we propose a data augmentation framework to systematically generate large amounts of effective training data. The proposed method enriches the training data which makes it contain sufficient variations of gestures and eases the pain of data collection.
The intuition behind the data augmentation is that DRAI representations vary with different gesture properties. After analyzing various practical scenarios of gesture executions, we summarize four factors which have significant influence on DRAI data under common sensing conditions, including distance to radar, angle of arrival, gesture speed, and trajectories of gestures. To improve the system performance when sensing at extreme angles, we further propose three techniques to augment data considering  the non-uniform angular resolution, variant radiated power and different geometric features of gesture trajectories.

\subsubsection {Different Distances}
Due to the fine-grained range information of DRAI, gestures at different distances lead to variations in DRAI. To measure the impact of the distance between radar and user, we perform push standing at 60cm and 80cm in front of the radar, respectively. Fig. \ref{push}(b) and (c) show the DRAI sequences, and we can observe vertical offset along range axis in DRAI sequences which results from different distances to the radar. Therefore, we can synthesize DRAIs of gestures performed at different distances by vertically translating all DRAIs in one sequence.    

\subsubsection {Different Angles}
To evaluate the impact of AoA, we perform pushing around the radar at different angles (i.e. 30\degree and -30\degree) with 80cm away from the radar. As illustrated in Fig. \ref{push}(c) and (d), we can observe that similarly to situations of different distances, variation of AoA results in a horizontal drift of DRAI. This is because the horizontal axis of DRAI represents angle information and the angular resolution of the radar is high enough to distinguish them. Therefore, DRAIs of gestures performed at different angles can be generated by horizontal translation. Note that translating DRAI horizontally is effective for small angular displacements, however, when sensing at extreme angles, more data augmentation methods are needed due to the characteristics of radar signals.

\subsubsection {Different Speeds}
Since the frame periodicity of the radar used in our system is 50ms, gestures lasting for 1 second will result in 20 consecutive DRAIs. It is clear that gesture samples with different speeds will have different lengths of produced DRAI sequences. Therefore, to simulate speed variations when users perform gestures, we can change the length of the DRAI sequence by downsampling and frame interpolation. To achieve this, we simply use linear frame interpolation that averagely mixes adjacent two frames to generate a new frame. 

\subsubsection {Different Trajectories}
In this work, we focus on in-air gestures that have different trajectories. For the simplicity of memory and execution, we design six pair-wise gestures which are push and pull, slide to left and slide to right, clockwise rotate, and anticlockwise rotate. Different pair-wise gestures have unique trajectories while the same pair-wise gestures have symmetry trajectories. Therefore, DRAI sequences can be reversed to produce their pair-wise gesture data which further increases the amount of data.

\subsubsection {Different Angular Resolution}
To achieve robust performance at extreme angles, the first consideration is that radar angular resolutions vary with angle of arrival $\theta$, which can be formulated as 
\begin{equation}\label{eq1}
	\theta_{res} = \frac{\lambda}{Nd\cos{\theta}} ,
\end{equation}
where $\lambda$ is the wavelength of the signal, $N$ is the number of receiving antennas and $d$ is the distance between adjacent receiving antennas. Therefore, performing gestures at different angles leads to inconsistent angular resolutions and simply translating along the angular domain cannot cope with this variation when gestures are performed at extreme angles. 
To mitigate the impact of non-uniform angular resolution, we process raw signals with different receiving antennas $N$ to produce augmented gesture samples with variant angular resolution. 

\subsubsection {Different Radiated Power}
Different from optical cameras, the received signal strength of the antenna is affected by the signal Angle of Arrival (AoA), which depends on the antenna radiation pattern. Specifically, to achieve longer sensing distance, most of the radiated power of the radar antenna is concentrated in a specific direction, which is termed as the main lobe. 
In contrast, the radiated power of other directions (i.e. outside the main lobe) is relatively lower~\cite{sidelobe}.  
Therefore, performing gestures at extreme angles (i.e. out of radar's main lobe) leads to a significant attenuation of signal strength. Specifically, since we aim to train a cross-angle classifier, we multiply DRAI samples collected at 0° with a scaling factor $\alpha$ to generate samples at different angles. 
Theoretically the scaling factor $\alpha$ of different angles should be proportional to the antenna gain. However, as the AoA of human hands is continuously changing when performing gestures, we experimentally fine-tune the $\alpha$ to better match the power distribution in practical scenarios. 

\begin{figure}[htbp]

		\begin{minipage}[t]{0.6\linewidth}
			\centering
			\includegraphics[scale= 0.15]{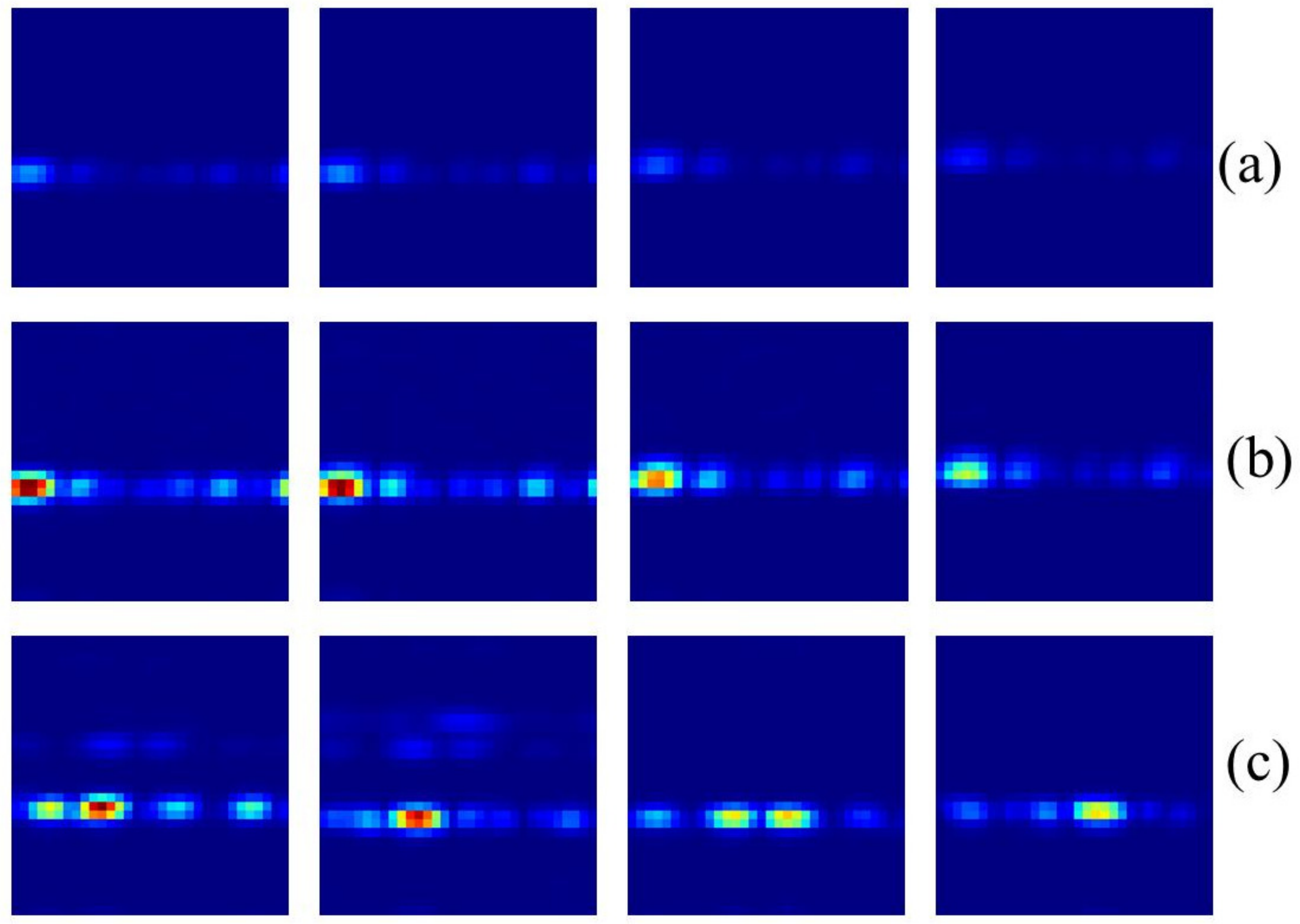}
			\caption{Left swipe at different angles. (a) raw DRAIs at 60°; (b) normalized DRAIs at 60°; (c) raw DRAIs at 0°.}
			\label{norm}
		\end{minipage}%
		\quad
		\begin{minipage}[t]{0.35\linewidth}
			\centering
			\includegraphics[scale= 0.115]{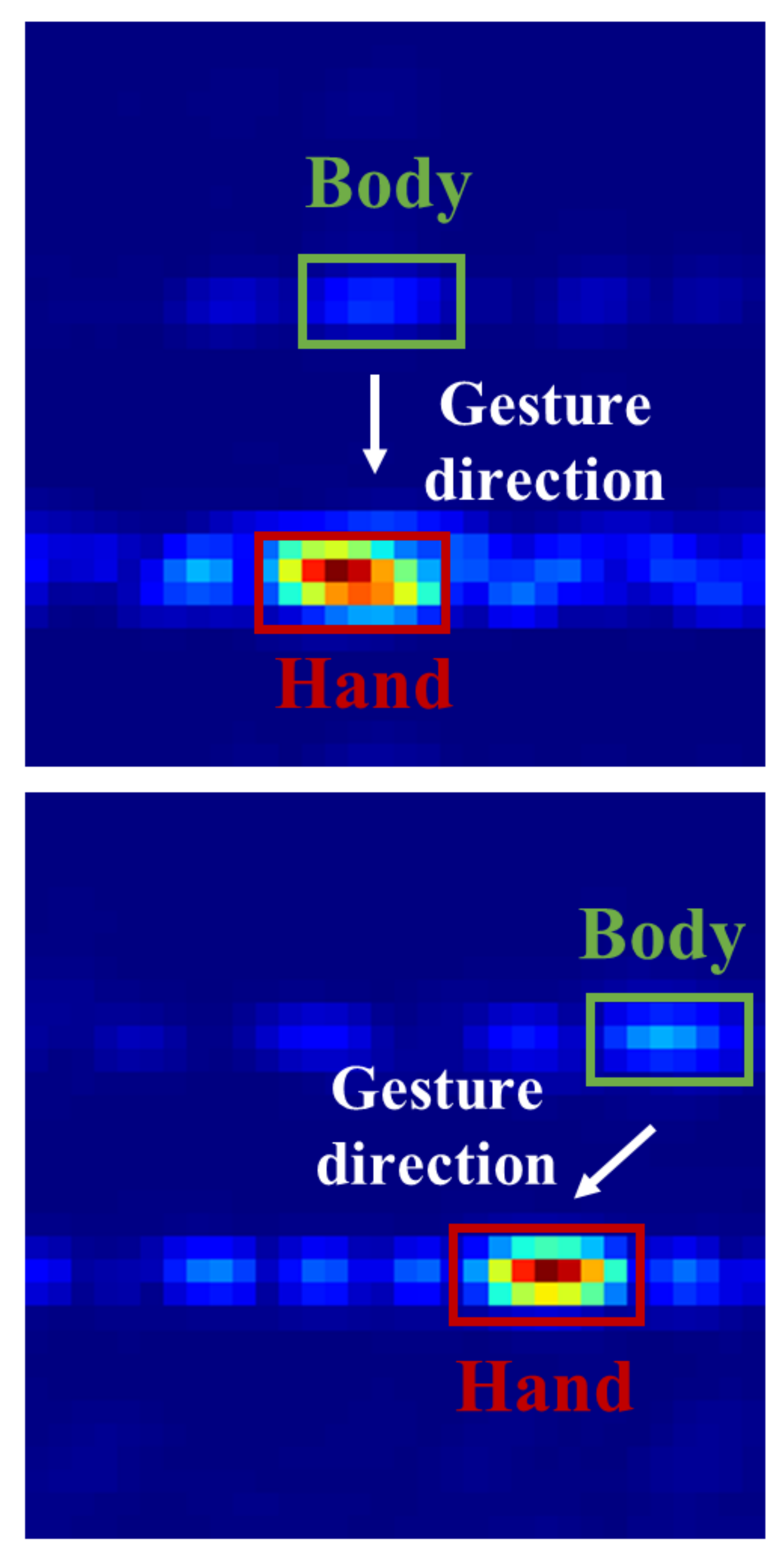}
			\caption{Push at 0° (up) and Push at 60° (down).}
			\label{direction}
		\end{minipage}%

\end{figure}

Fig. \ref{norm} (a) shows a raw DRAI sequence of left swipe at 60° and Fig. \ref{norm} (b) shows the corresponding normalized DRAI sequence.  We can observe that the signal strength of Fig. \ref{norm} (a) is apparently weaker than that of Fig. \ref{norm} (c) and their signal patterns become more similar after normalization (i.e. the signal strength of Fig. \ref{norm} (b) is consistent with Fig. \ref{norm} (c)).

\subsubsection {Different Geometric Features}
Another key observation is that when gestures are performed at extreme angles, geometric features (i.e. directions or scales) of gesture trajectories will be deformed. For example, as shown in Fig. \ref{norm}, the horizontal movement of the brightest spot in Fig. \ref{norm} (c) is clearly larger than that of Fig. \ref{norm} (b). This is because the radar angular resolution decreases with the increase of angular displacement, which leads to less movement along the horizontal axis in DRAI. 
Besides, the direction of gesture trajectory also changes when performing gestures at extreme angles, as shown in Fig. \ref{direction}.

\begin{figure}[htbp]
	\begin{center}
		\includegraphics[scale=0.115]{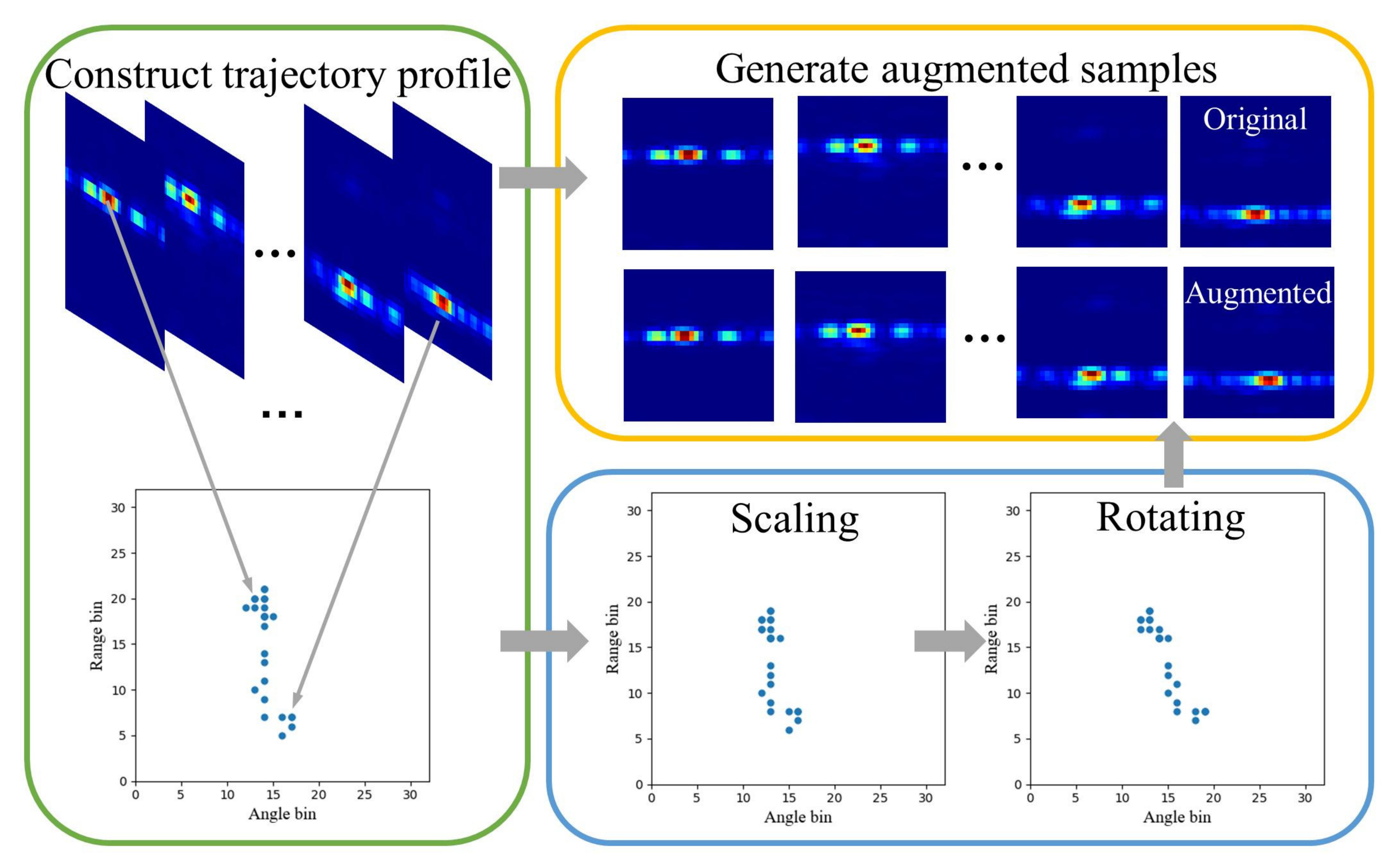}
		\caption{Procedure of data augmentation for different geometric features.}
		\label{rotate}
	\end{center}
\end{figure}
To compensate for the different geometric features, we utilize geometric transformations including rotating and scaling to generate gesture samples with different directions and scales. Note that the rotating and scaling should be applied on the gesture trajectory of DRAI sequences (i.e. spatial-temporal domain) rather than a single frame of DRAI (i.e. spatial domain).
Therefore, we propose trajectory profile  $\mathscr{P}$ to extract geometric features of gesture trajectories and map the DRAI sequences to point sets. As shown in Fig. \ref{rotate}, we firstly represent the hand position in frame $t$ with the coordinate  $(x_t,y_t)$ of the pixel with maximum magnitude. Then, the trajectory profile $\mathscr{P}$ can be constructed with a set of points, which is formulated as $\mathscr{P} = \{(x_t,y_t)| t=1,2,...,T\}$ and $T$ is the length of the DRAI sequence.  

Next, we apply rotating and scaling transformation on the original trajectory profile to obtain its augmented version. The rotating transformation can be expressed as 
{\small
\begin{equation}
	\left[ \begin{array}{c}
		x' \\ y' \\ 1  
	\end{array}
	\right] \!=\!
	\left[ 
	\setlength{\arraycolsep}{1.5pt}
	\begin{array}{ccc}
		\cos{\beta}& -\sin{\beta} & r_x(1-\cos{\beta})+r_y\sin{\beta} \\
		\sin{\beta}& \cos{\beta} & r_y(1-\cos{\beta})-r_x\sin{\beta} \\
		0 & 0 & 1
	\end{array}
	\right]
	\times
	\left[ \begin{array}{c}
      		x \\ y \\ 1  
	\end{array}
	\right] 
\end{equation}
}
where $\beta$ is the rotating angle and $(r_x,r_y)$ is the rotating center. We choose the point with the maximum Euclidean distance which represents the position of human body as the rotating center, in order to simulate variations of user orientations and gesture directions. The rotating center $(r_x,r_y)$ can be calculated by
\begin{equation}
	(r_x,r_y) = \mathop {\max}_{(x_t,y_t)\in \mathscr{P}} x_t^2+y_t^2
\end{equation}
The scaling transformation on trajectory profiles can be described as

\begin{equation}
	\left[
	\begin{array}{c}
		x' \\ y' \\ 1  
	\end{array}
	\right] 
	=
	\left[ \begin{array}{ccc}
		\gamma_x & 0 & s_x(1-\gamma_x) \\
		0 & \gamma_y & s_y(1-\gamma_y) \\
		0 & 0 & 1
	\end{array}
	\right]
	\times
	\left[ \begin{array}{c}
		x \\ y \\ 1  
	\end{array}
	\right] 
\end{equation}
where $\gamma_x$ and $\gamma_y$ is the scaling factor of the $x$ axis and $y$ axis, respectively. $(s_x,s_y)$ is the scaling center which is computed as 

\begin{equation}
	s_x = \frac{1}{T}\sum_{t=1}^T x_t,
	s_y = \frac{1}{T}\sum_{t=1}^T y_t, (x_t,y_t)\in \mathscr{P}
\end{equation}

After transforming the trajectory profile, we compute the offset $(\Delta x_t,\Delta y_t) = (x'_t,y'_t)-(x_t,y_t) $ of frame $t$ in the DRAI sequence and translate the original frame according to the offset. Different from translating the whole DRAI sequence with the same displacement to simulate variations of locations, the transformations mentioned above can shift each frame in the DRAI sequence with a specific offset, thus generating gesture samples with different geometric features.  

\begin{figure*}[t]
	\begin{center}
		\includegraphics[scale=0.17]{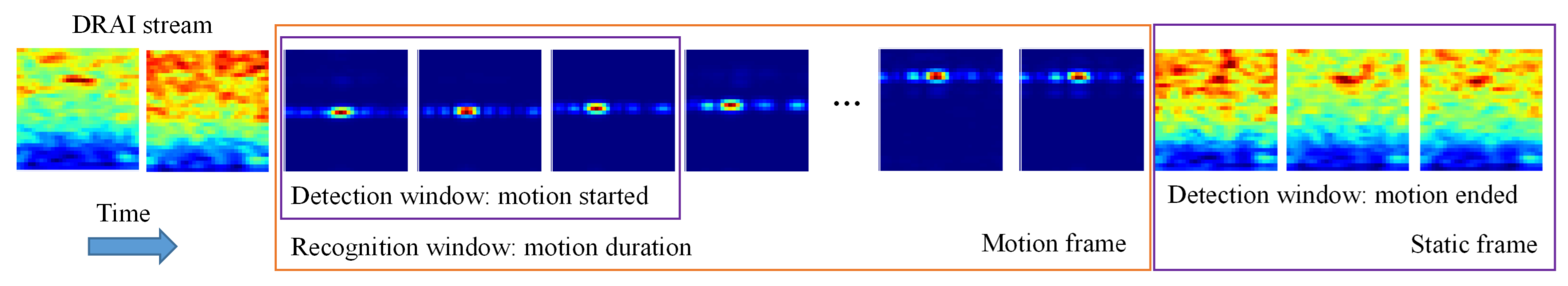}
		\caption{Dynamic window mechanism for gesture segmentation.}
		\label{window}
	\end{center}
\end{figure*}

\subsection{Gesture Segmentation}
To make the system work in real-time and multi-person scenarios\cite{mid,multitarget}, we design a spatial-temporal segmentation algorithm based on CFAR \cite{CFAR} and CLEAN \cite{clean} to separate signals of multiple users in the spatial domain, then detect gesture boundaries of DRAI sequences in the temporal domain. Specifically, the first step of spatial segmentation is to perform angle FFT on the zero-frequency component of range doppler images to obtain Static Range Angle Image (SRAI) and filter out interfering users with large movements. Then, an iterative detection and cancellation strategy \cite{clean,CFAR} is used to detect multiple static targets in the SRAI.  
After that, the closer detected target is chosen as the intended user and a fixed region of interest is set to prevent the influence of interfering users.

To overcome the limitation of the fixed-length window, we propose a dynamic window mechanism to adjust the window size automatically. 
To be specific, the first step is to distinguish whether the current frame is a motion frame (i.e. human body movement occurs in the detection range of the radar) or a static frame (i.e. no moving object). Then, a detection window is sliding along the DRAI stream to detect motion boundaries as shown in Fig. \ref{window}. When all frames inside the detection window are labeled as motion frames for the first time, it will be considered as the starting of a hand gesture or other unexpected motions. Once a motion starting is detected, the size of the recognition window begins to increase until all frames inside the detection window are labeled as static frames, in other words, the motion ends. After that, frames belonging to the recognition window are passed into the classifier to decide whether it is a predefined gesture or not.

\begin{figure}[htbp]
	\subfigure[Static frame]{
		\begin{minipage}[t]{0.5\linewidth}
			\centering
			\includegraphics[scale= 0.07]{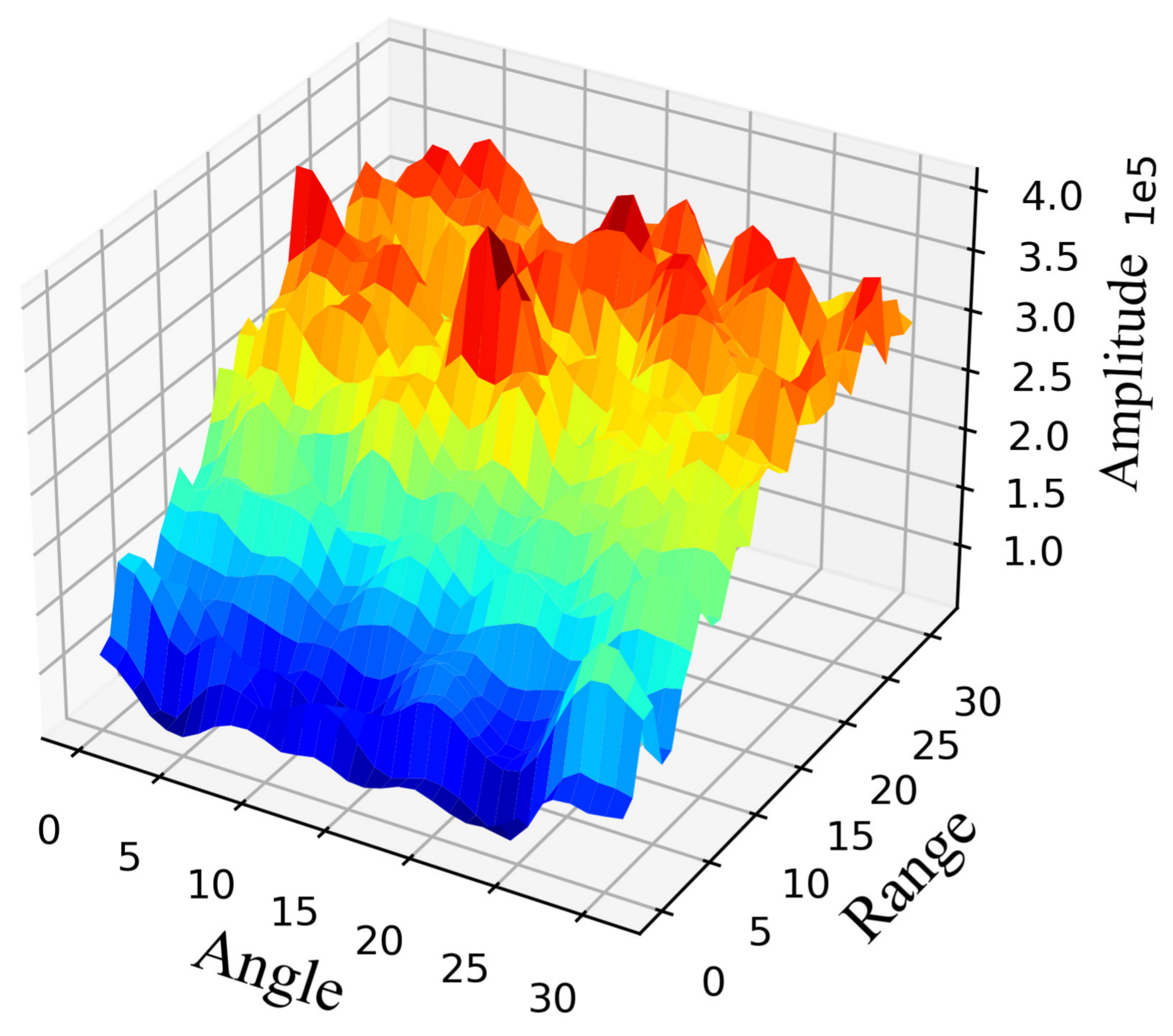}
			%\caption{fig1}
		\end{minipage}%
	}%
	\subfigure[Motion frame]{
		\begin{minipage}[t]{0.5\linewidth}
			\centering
			\includegraphics[scale= 0.07]{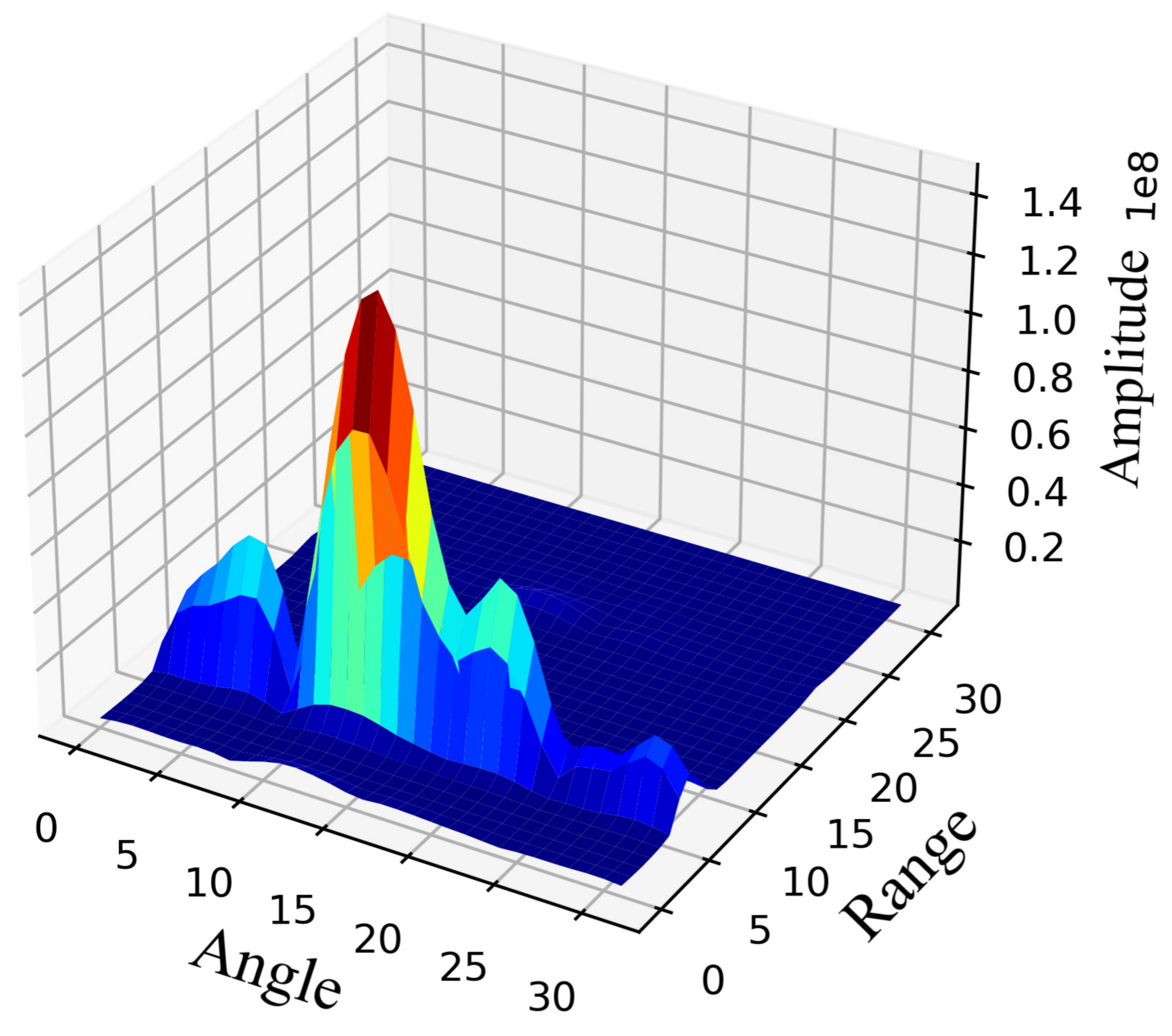}
			%\caption{fig2}
		\end{minipage}%
	}%
	
	\caption{ Difference between the motion frame and static frame.}
	\label{difference}
\end{figure}

The classification of motion frame is based on the fact that energy distributions of DRAI show a great difference in different situations, as shown in Fig.  \ref{difference}. 
%Specifically, there is always a series of peaks in motion frame, while the static frame looks like undulating hills. 
Specifically, there is always a series of explicit patterns (i.e. multiple high-intensity peaks) in motion frames caused by gestures, while static frames are perturbed by random noise severely. The larger differences between the energy of the peak cell and the energy of the background noise, the more likely it would be a motion frame. 

Assuming that all cells $(x,y)$ of the DRAI is set $\mathscr{C}$ and the signal energy of cell $(x,y)$ is $E(x,y)$, the energy of the peak cell is calculated as 
\begin{equation}
	E_{peak} =  \mathop{max}\limits_{(x,y) \in \mathscr{C}} E(x,y)
\end{equation}
After finding the peak cell, we estimate the average energy of background noise as  
\begin{equation}
	E_{noise} = \mathop{mean}\limits_{(x,y) \in \mathscr{B}} E(x,y)
\end{equation}
where 	$\mathscr{B} = \{(x,y)|x \in \left[1,x_p-2\right) \cup  \left(x_p+2,32\right] ,y \in \left[1,y_p-4\right) \cup \left(y_p+4,32\right] \}$ and $(x_p,y_p)$ is the location of the peak cell.
The current frame in the DRAI sequence will be marked
as a motion frame if the motion indicator $\eta$ is greater than a preset threshold  $T_{motion}$:
%The current frame in SRAI sequence will be marked as motion frame if the following equation is satisfied
\begin{equation}
  \eta = log(\frac{E_{peak}+E_{noise}}{E_{noise}}) > T_{motion},
\end{equation}

\begin{figure}[htbp]
	\begin{center}
		\includegraphics[scale=0.06]{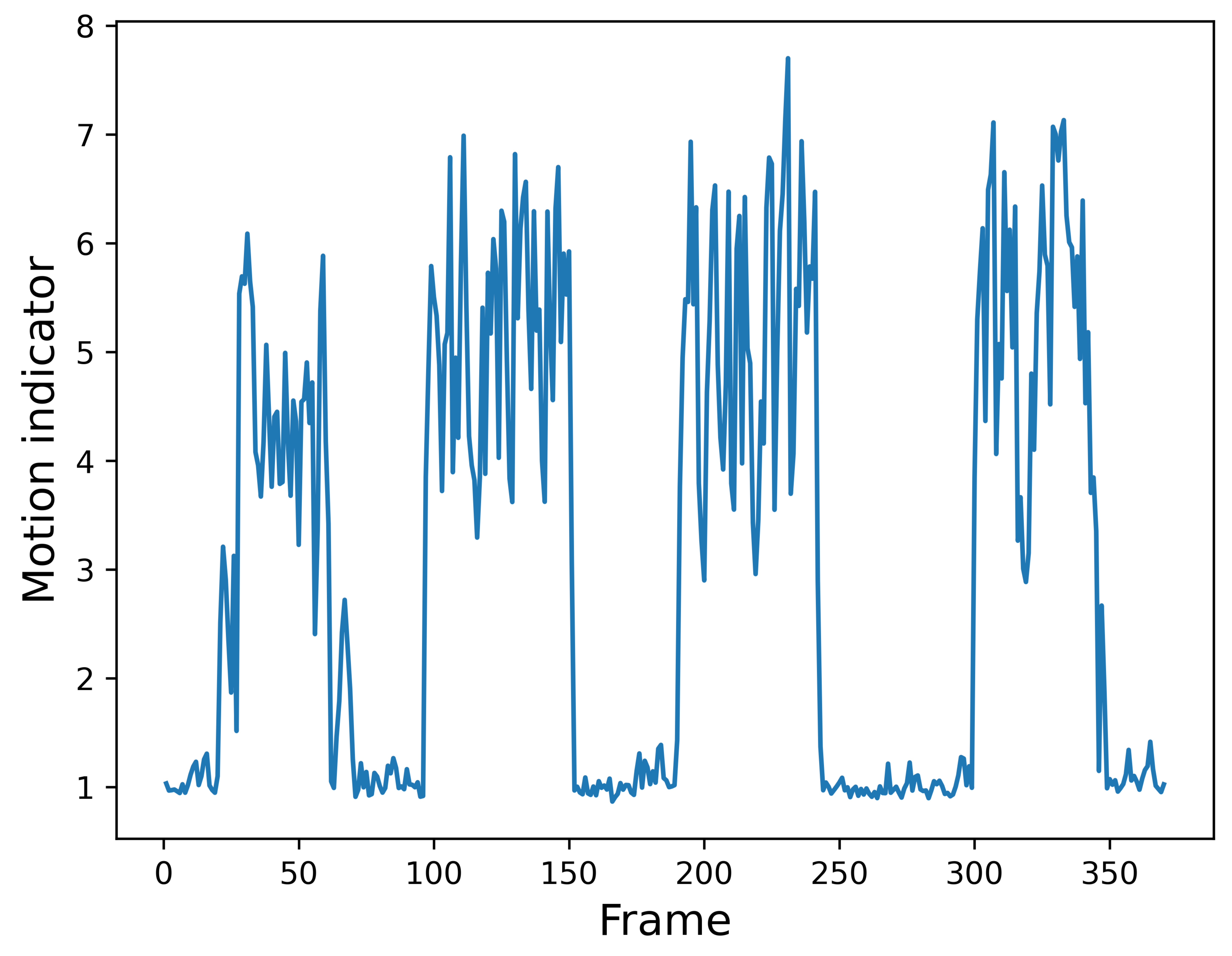}
		\caption{Variations of motion indicator when gestures are performed continuously.}
		\label{segment}
	\end{center}
\end{figure}

As shown in Fig. \ref{segment}, the motion indicator drastically increases to high values when the user is performing gestures, while staying stable at low values when the user is static. In our experiment, we set the threshold $T_{motion}$ as 1.8.

\subsection{Gesture Recognition}

The consecutive DRAIs describe how the doppler power distribution changes corresponding to a particular kind of gesture. In order to successfully recognize gestures, we need to extract spatial-temporal features of DRAIs. Recently, Convolutional Neural Network (CNN) \cite{cnn} exhibits significant progress in the field of computer vision in learning spatial features from images automatically while Long Short-Term Memory (LSTM) \cite{lstm} has demonstrated great power in modeling temporal information from time-series data. Therefore, we design a neural network consisting of a frame model which employs CNN to extract spatial features from each single DRAI and a sequence model which utilizes LSTM to learn temporal dependencies of the entire DRAI sequence.

\begin{figure}[htbp]
\begin{center}
	\includegraphics[scale=0.14]{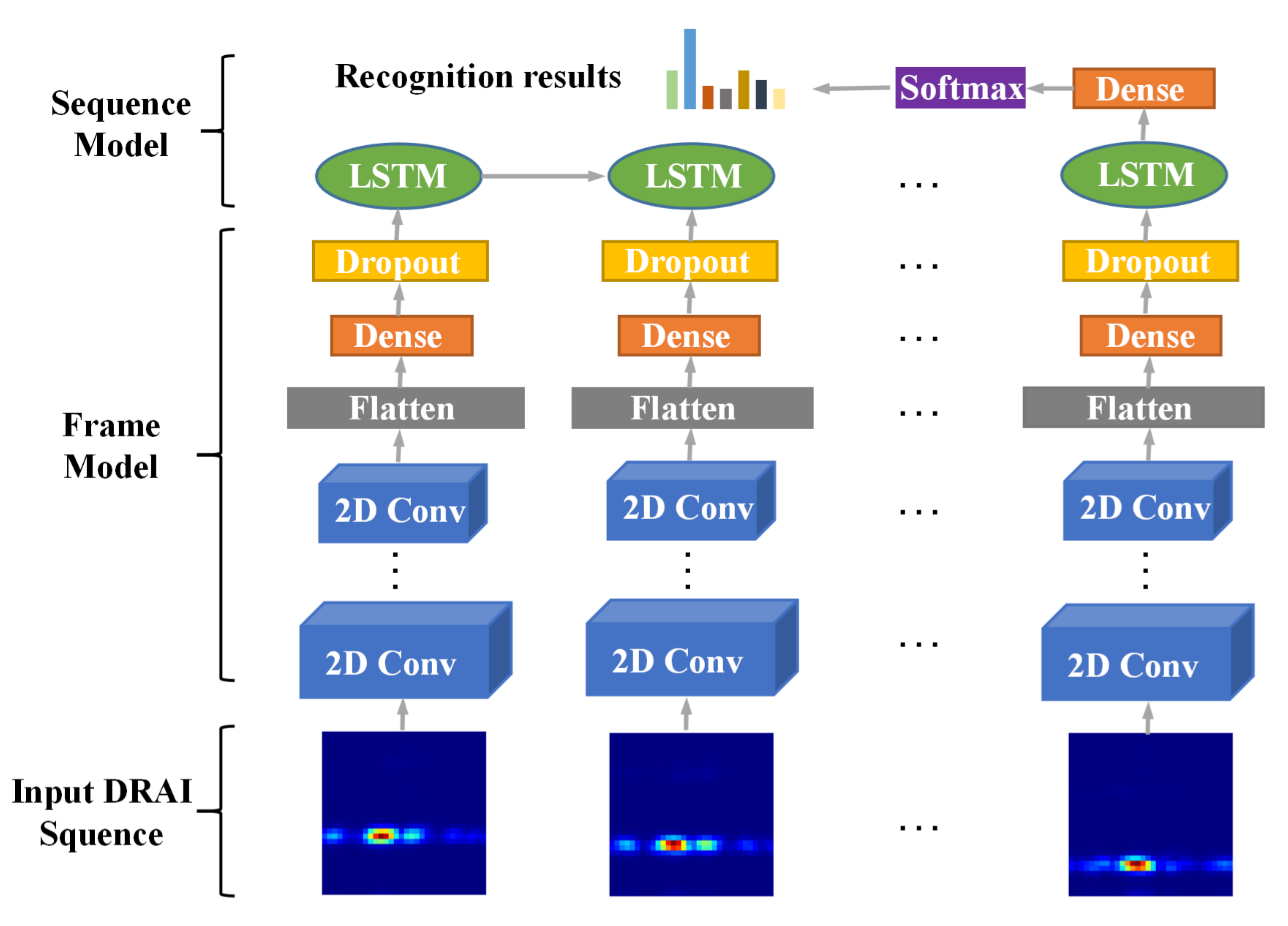}
	\caption{Network architecture. The frame model employs CNN for spatial features extraction and the sequence model utilizes LSTM for temporal modeling.}
	\label{network}
\end{center}
\end{figure}

To better match data characteristics of DRAIs and reduce computational consumptions, we design a shallow neural network which  has three convolution layers and one LSTM layer and can run on CPU-only devices in real-time. The last time step of LSTM output is passed into a fully connected layer and a softmax layer to perform gesture recognition. The loss function of our model can be expressed as
\begin{equation}
	\begin{aligned}
		Loss(Y,c) &= -log(\frac{exp(Y[c])}{\sum_{j=0}^{C-1} exp(Y[j])}) \\ 
				  &= -Y[c] + log(\sum_{j=0}^{C-1} exp(Y[j]))
	\end{aligned}
\end{equation} 
where $Y$ refers to the output of the last fully connected layer, and $C$ is the number of gesture classes.

\section{Implementation}

\subsection{Dataset}

\begin{figure*}[t]
	\begin{center}
		\includegraphics[scale=0.16]{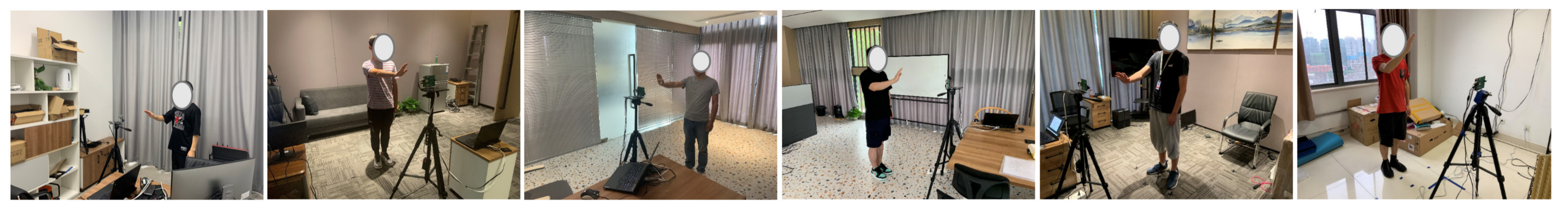}
		\caption{Different indoor environments in our mmWave gesture dataset.}
		\label{rooms}
	\end{center}
\end{figure*}

To evaluate the performance of DI-Gesture, we collect gesture data from 25 volunteers, 6 environments, and 5 locations. Fig. \ref{rooms} shows different indoor environments including the living room, meeting room, bedroom, laboratory, and two office rooms. Each environment has different sizes and furniture placement, which results in different multipath effects. We select six common gestures which are easy to memorize and execute as predefined gestures. We also collect other human actions as negative samples to improve the robustness of the classifier and filter unexpected motions in real-time application scenarios. The distances and angles of location 1-5 are (0.6m, 0\degree), (0.8m, 0\degree), (1.0m, 0\degree), (0.8m, -30\degree) and (0.8m, 30\degree), respectively. Each volunteer is asked to perform each kind of gesture 5 or 10 times at each location. The data collecting process has spent 15 days. In total, we have collected 24050 samples, consisting of 10650 gesture samples and 13400 negative samples. The detailed description of our dataset is shown in Table. \ref{dataset}.

\begin{table*}[htbp]
	\centering
	\caption{Detailed description of DI-Gesture dataset.}
	\label{dataset}
	\resizebox{\linewidth}{!}{
	\begin{tabular}{|c|l|c|c|}
		\hline
		\multicolumn{2}{|c|}{\multirow{2}[2]{*}{Predefined Gestures}} & \multicolumn{2}{c|}{\multirow{2}[2]{*}{Push (PH), Pull (PL), Left swipe (LS) Right swipe (RS), Clockwise turning (CT) and Anticlockwise turning (AT)}} \\
		\multicolumn{2}{|c|}{} & \multicolumn{2}{c|}{} \\
		\hline
		\multicolumn{2}{|c|}{\multirow{2}[2]{*}{Negative Samples}} & \multicolumn{2}{c|}{\multirow{2}[2]{*}{Lifting  right arm,  Lifting  left arm, Sitting down, Standing up, Waving hands, Turning around, Walking (NG)}} \\
		\multicolumn{2}{|c|}{} & \multicolumn{2}{c|}{} \\
		\hline
		\multirow{12}[12]{*}{ Users (25)} & \multicolumn{1}{c|}{\multirow{2}[2]{*}{User A-User G (7)}} & \multicolumn{2}{c|}{\multirow{2}[2]{*}{7 Users x 5 Rooms x  5 Locations x (6 Gestures  x 5 Instances + 60 Negative samples) = 12250 Samples}} \\
		&       & \multicolumn{2}{c|}{} \\
		\cline{2-4}          & \multicolumn{1}{c|}{\multirow{2}[2]{*}{User H-User I (2)}} & \multicolumn{2}{c|}{\multirow{2}[2]{*}{2 Users x 4 Rooms x  5 Locations x (6 Gestures x 5 Instances + 60 Negative samples) = 2800 Samples}} \\
		&       & \multicolumn{2}{c|}{} \\
		\cline{2-4}          & \multicolumn{1}{c|}{\multirow{2}[2]{*}{User J-User L (3)}} & \multicolumn{2}{c|}{\multirow{2}[2]{*}{3 Users x 3 Rooms x 5 Locations x  (6 Gestures x 5 Instances + 60 Negative samples) = 3150 Samples}} \\
		&       & \multicolumn{2}{c|}{} \\
		\cline{2-4}          & \multicolumn{1}{c|}{\multirow{2}[2]{*}{User M-User N (2)}} & \multicolumn{2}{c|}{\multirow{2}[2]{*}{2 Users x 2 Rooms x 5 Locations x (6 Gestures x 5 Instances + 60 Negative samples) = 1400 Samples}} \\
		&       & \multicolumn{2}{c|}{} \\
		\cline{2-4}          & \multicolumn{1}{c|}{\multirow{2}[2]{*}{User O-User R (4)}} & \multicolumn{2}{c|}{\multirow{2}[2]{*}{4 Users x 1 Room x 5 Locations x (6 Gestures  x 10 Instances + 60 Negative samples) = 2000 Samples}} \\
		&       & \multicolumn{2}{c|}{} \\
		\cline{2-4}          & \multicolumn{1}{c|}{\multirow{2}[2]{*}{User S-User Y (7)}} & \multicolumn{2}{c|}{\multirow{2}[2]{*}{7 Users x 1 Room x 5 Locations x (6 Gestures x 5 Instances + 60 Negative samples) = 2450 Samples}} \\
		&       & \multicolumn{2}{c|}{} \\
		\hline
		\multirow{1}[1]{*}{Rooms (6)} & \multicolumn{1}{l|}{ \makecell[c]{ Meeting room: 4900 Samples \\ Living room: 5250 Samples \\ Bedroom: 4550 Samples \\ Laboratory: 3050 Samples \\ 2 office rooms: 6300 Samples}} & \multirow{1}[2]{*}{\textbf{Total Domains and Samples}} & \multicolumn{1}{c|}{\multirow{1}[2]{*}{\makecell[c] {6 environments x 25 users x 5 locations = 750 domains \\ 10650 Gesture Samples + 13400 Negative Samples = 24050 Samples}}} \\

		\hline
	\end{tabular}%
}
	\label{tab:addlabel}%
\end{table*}%

\subsection{Device Configuration}

We have implemented our gesture recognition system using TI AWR1843 mmWave radar and DCA1000 real-time data acquisition board. Each radar frame has 128 chirps and each chirp has 128 sample points. The frame rate, range resolution, and velocity resolution of radar are 20fps, 0.047m, and 0.039m/s, respectively. We activate 2 transmitting antennas and 4 receiving antennas to obtain an approximately angular resolution of 15\degree. We only keep the first 32 range bins (i.e. 1.5m) in DRAI in order to eliminate background noise and reduce the computational cost of neural networks. The Angle-FFT size is set as 32 to improve the angle resolution. Therefore, the size of DRAI is 32 x 32.

\subsection{Network Implementation}
The frame model has 3 convolutional layers with kernel size 3x3, 1 fully connected layer with 128 units and batch normalization. The number of filters of the three convolutional layers increases from 8, 16 to 32. The sequence model consists of 1 LSTM layer with a size of 128 hidden units and 1 fully connected layer with 128 inputs to obtain gesture probability. We set the activation function as ReLU and the dropout rate as 0.5. The network is trained with Adam optimizer with a learning rate of 0.0001 and a batch size of 128. The training epoch of models without data augmentation is 100 to prevent overfitting while extending to 200 when trained with data augmentation to make the augmented data cover as many variations as possible.

\subsection{Data augmentation}
Suppose $U$ is a uniform distribution, the hyperparameters of the data augmentation framework are set as $\delta_x\sim{U(-6,6)}$, $\delta_y \sim U(-20,20)$, $\delta_k \sim U(3,5)$, $\beta \sim U(-\frac{\pi}{12},\frac{\pi}{12})$, and $\gamma \sim U(0.8,1.2)$, which represent translation along range axis, translation along angle axis, the interval of inserting or removing a frame, the rotating angle, and the scaling factor of gesture trajectories, respectively.  The normalizing factor of DRAI magnitude is set as $\alpha \sim U(0.4,1.0) $ at 45° and $\alpha \sim U(0.2,0.8) $ at 60° according to the antenna pattern of AWR1843.

\section{Evaluation}
\label{performance}
In this section, we first evaluate the in-domain performance of DI-Gesture, then we evaluate the recognition ability of DI-Gesture under different domain factors, including new users, new environments, and new locations. After that, we make comparisons with the state-of-the-art \cite{radarnet} and make a detailed analysis of the proposed data augmentation framework. Finally, we evaluate the performance of real-time recognition and the computational consumption of our system.

\subsection{Ability of In-Domain Recognition}
\begin{figure}[htbp]
	%\subfigure[RadarNet (86.15\%)]{
	\begin{minipage}[t]{0.23\textwidth}
		\centering
		\includegraphics[scale= 0.055]{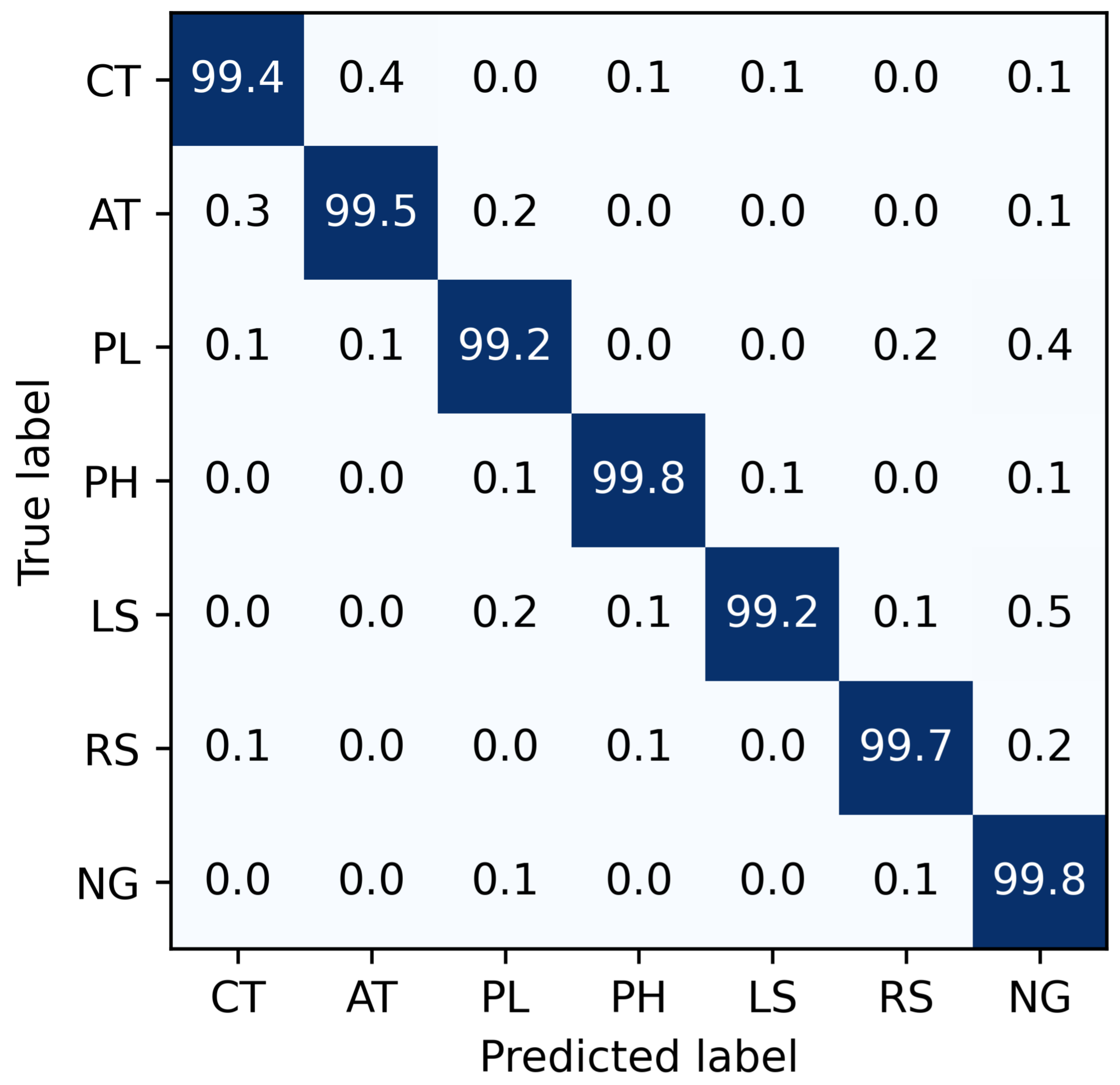}
		\caption{Confusion matrix of in-domain test.}
		\label{indomainconf}
	\end{minipage}%
	\quad
	\begin{minipage}[t]{0.23\textwidth}
		\centering
		\includegraphics[scale= 0.055]{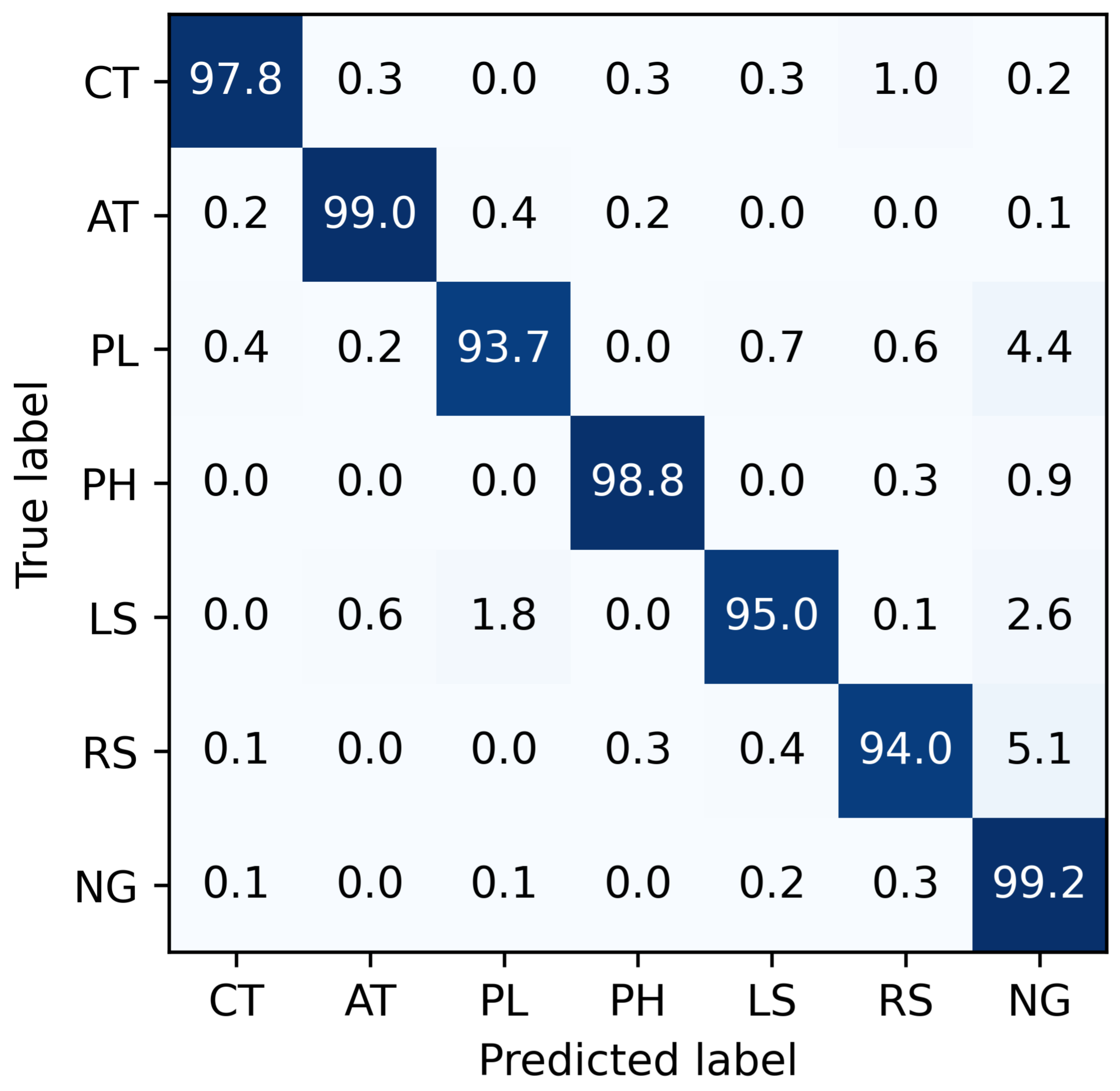}
		\caption{Confusion matrix of new user test.}
		\label{newuserconf}
	\end{minipage}%
\end{figure}

In-domain recognition means that the training data and the testing data belong to the same domains. We take 80\% of data from each domain for training and then test the remaining 20\% with 5-fold cross-validation. The average accuracy of in-domain test reaches over 99\%. To obtain a clearer view of the result, we present the confusion matrix of the in-domain test in Fig. \ref{indomainconf}, from which we can observe that all gestures can be recognized with high accuracy and non-gestures can also be classified well. This is mainly because training data and test data from the same domains share familiar characteristics. The result also demonstrates that DRAI indeed carries inherent characteristics of gesture which can be effectively extracted by the proposed neural network.

\subsection{Ability of Cross-Domain Recognition}
\emph{(1) Person Variety.} Different individuals may perform gestures with diversified speeds and scales, which could impact DRAI features. To evaluate the person-independent performance of DI-Gesture, we train DI-Gesture with gesture data from User A-G and test with the data of the remaining 18 persons. As a result, DI-Gesture achieves an accuracy of 97.92\% for different users. The confusion matrix of the new user test is shown in Fig. \ref{newuserconf}. We observe that the accuracy of all gestures reaches over 90\%, Compared with other gestures, gestures "pull", "left swipe" and "right swipe" are more likely to be mis-classified with negative samples. This is because there are some undefined actions which are very similar to these three gestures, such as lifting arms and waving hands. Overall, DI-Gesture is able to accurately recognize all gestures for new users.

\begin{figure}[htbp]
	%\subfigure[RadarNet (86.15\%)]{
		\begin{minipage}[t]{0.23\textwidth}
			\centering
			\includegraphics[scale= 0.052]{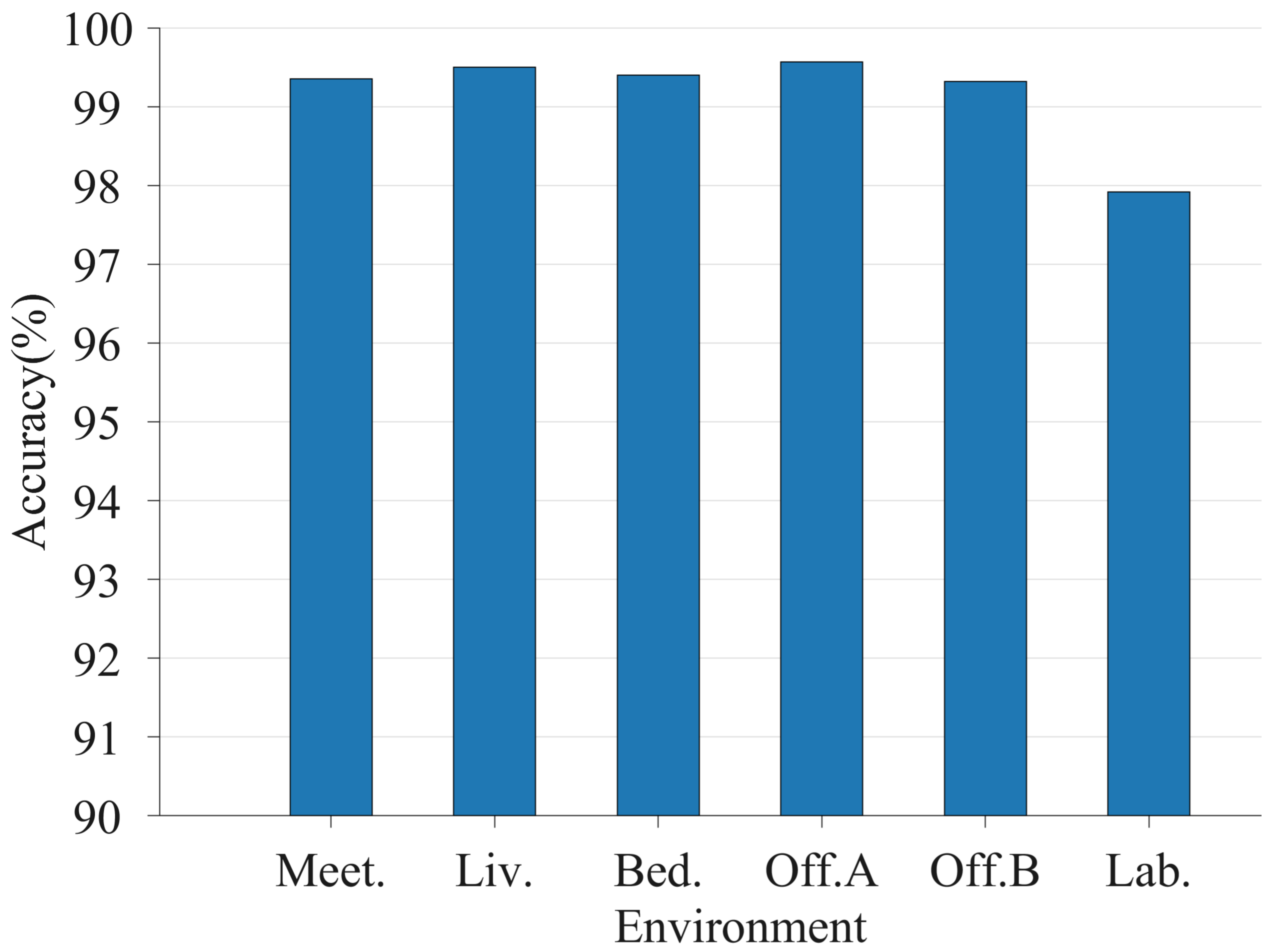}
			\caption{Accuracy of new environment test.}
			\label{newenv}
		\end{minipage}%
	\quad
		\begin{minipage}[t]{0.23\textwidth}
			\centering
			\includegraphics[scale= 0.052]{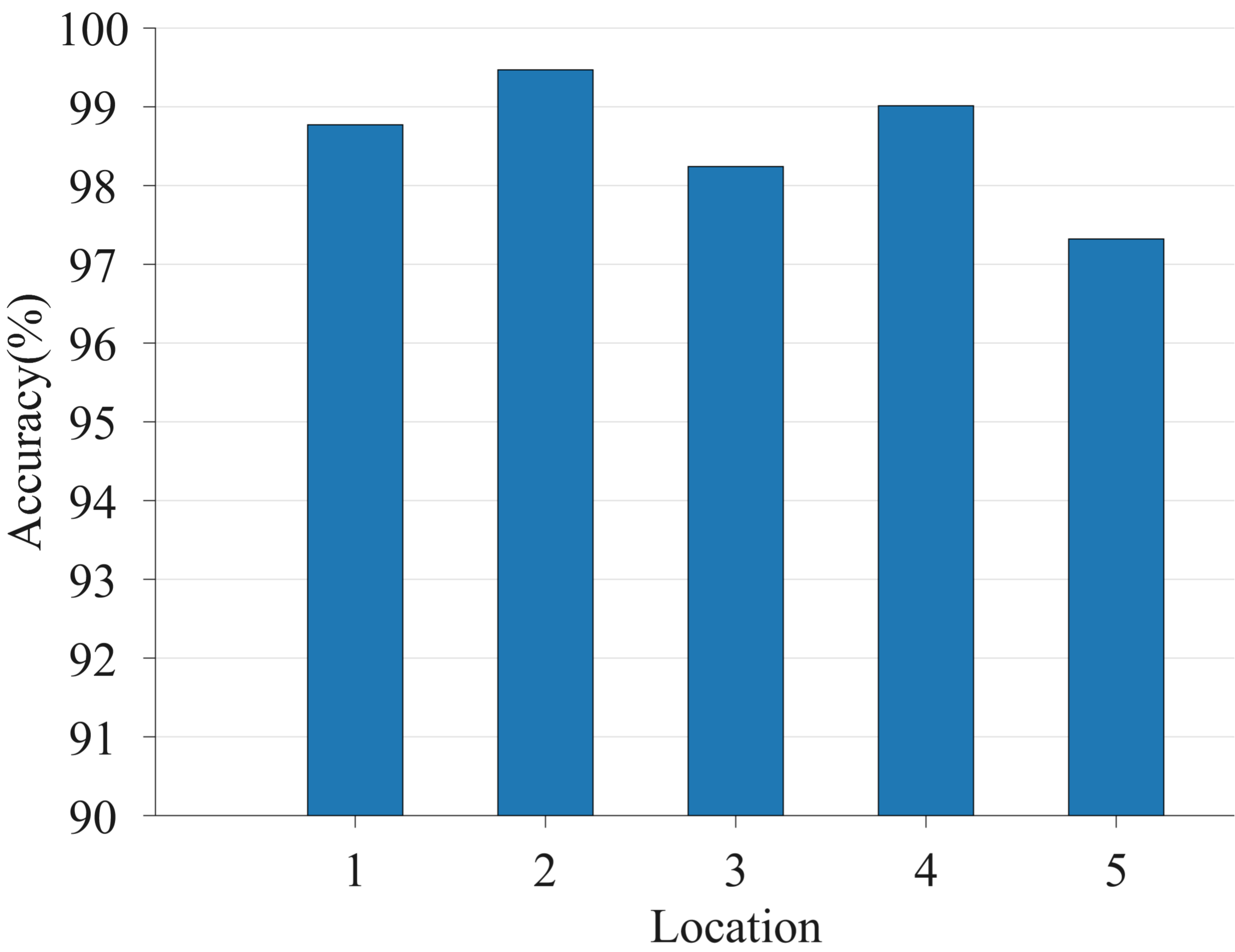}
			\caption{Accuracy of new location test.}
			\label{newloc}
		\end{minipage}%
	\label{envloc}
	
\end{figure}

\emph{(2) Environment diversity.} Environment plays an important role in wireless sensing systems due to different multipaths effect caused by different room sizes, furniture placement and device deployment. To investigate the cross-environment performance of DI-Gesture, we adopt leave-one-environment-out test meaning that taking data collected in 1 room as the test set and the other 5 rooms as the training set. As Fig. \ref{newenv} depicts, the average accuracy of the new environment test is 99.18\%, which indicates the strong robustness of DI-Gesture across different environments. Note that users performing gestures at the laboratory are completely different from the rest rooms, which is a more challenging scenario. As a result, DI-Gesture can still achieve an accuracy of 97.92\%.

\emph{(3) Location variation.} To investigate the performance of DI-Gesture at different locations, we conduct leave-one-location-out test, which denotes one location as the test set and the remaining four locations as the training set. As shown in Fig. \ref{newloc}, our system achieves promising results at different locations with an average accuracy of 98.56\%. In conclusion, DI-Gesture is robust against location variation.

\begin{table*}[t]
	\centering
	\caption{Comparison of computational cost, in-domain, and cross-domain recognition accuracy.}
	\label{comparison}
	%\resizebox{\linewidth}{!}{
	\setlength{\tabcolsep}{4mm}{
	\begin{tabular}{ccccccc}
		\toprule
		\textbf{Model} & \textbf{FLOPs [M]} & \textbf{Params [K]} & \textbf{In-domain} & \textbf{Cross user} & \textbf{Cross environment} & \textbf{Cross location} \\
		\midrule
		RadarNet \cite{radarnet} & 0.98  & 37.29 & 97.80\% & 88.64\% & 92.16\% & 83.24\% \\
		DI-Gesture-Lite & \textbf{0.93} & \textbf{40.79} & \textbf{99.11\%} & \textbf{97.05\%} & \textbf{98.47\%} & \textbf{97.51\%} \\
		\bottomrule
	\end{tabular}%
}
	\label{tab:addlabel}%
\end{table*}%

\subsection{Comparison with the state-of-the-art}
To demonstrate the superior domain-independent recognition ability of DI-Gesture, we implement RadarNet proposed in \cite{radarnet} with the same TI-AWR1843 board for comparison. Since RadarNet is an efficient neural network developed for mobile devices, we propose DI-Gesture-Lite with a comparable model size to guarantee fairness. Specifically, we reduce the CNN embedding vector size from 128 to 32 and decrease the number of LSTM hidden nodes from 128 to 64, while keeping other network structures unchanged to obtain a new model, denoted as DI-Gesture-Lite. The floating point operations (FLOPs) and the number of model parameters of RadarNet and DI-Gesture-Lite are shown in Table. \ref{comparison}. 

\begin{figure*}[htbp]
	\begin{minipage}[t]{0.33\textwidth}
		\centering
		\includegraphics[scale= 0.065]{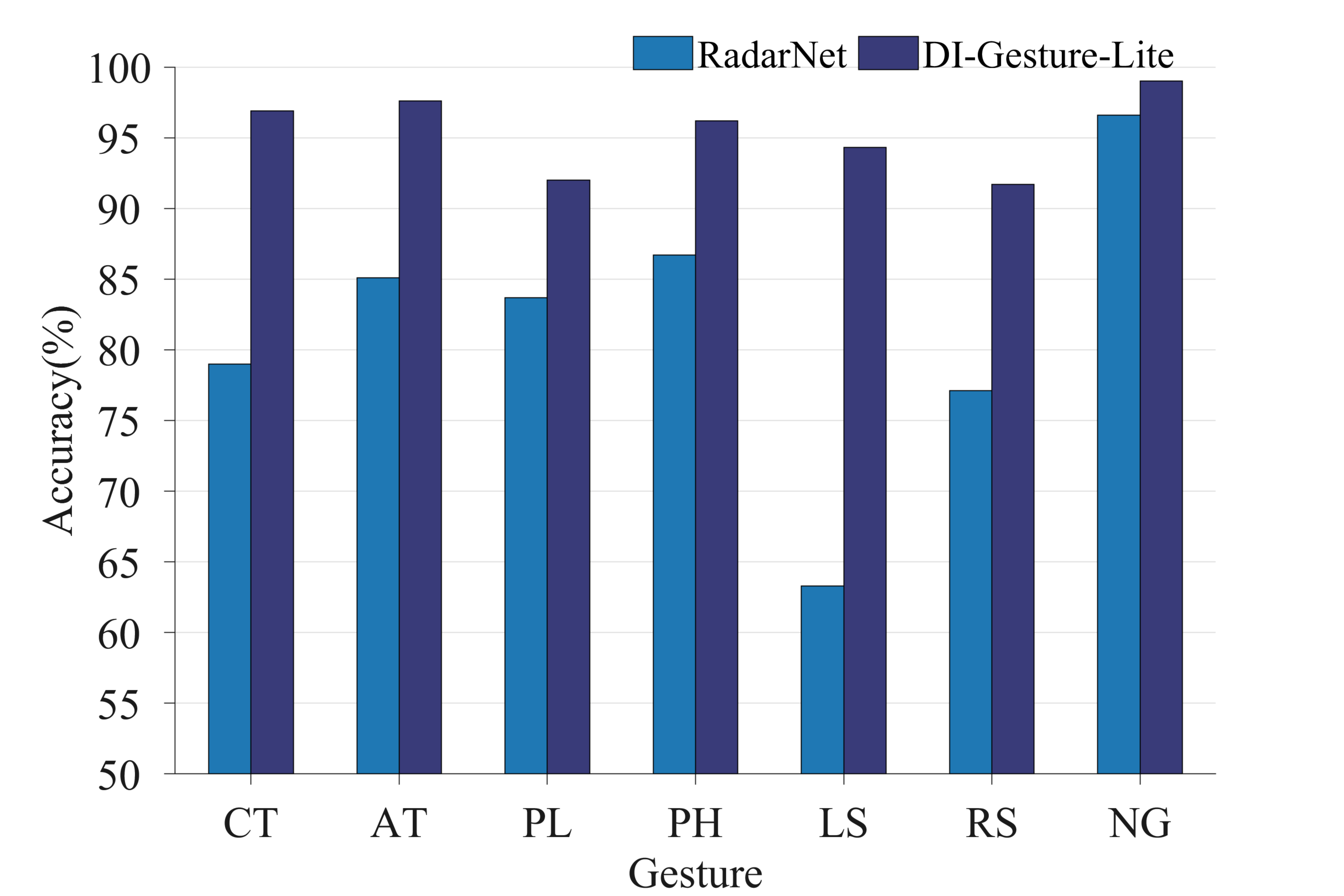}
		\caption{Comparison of new user test.}
		\label{comparenewuser}
	\end{minipage}%
	\begin{minipage}[t]{0.33\textwidth}
		\centering
		\includegraphics[scale= 0.065]{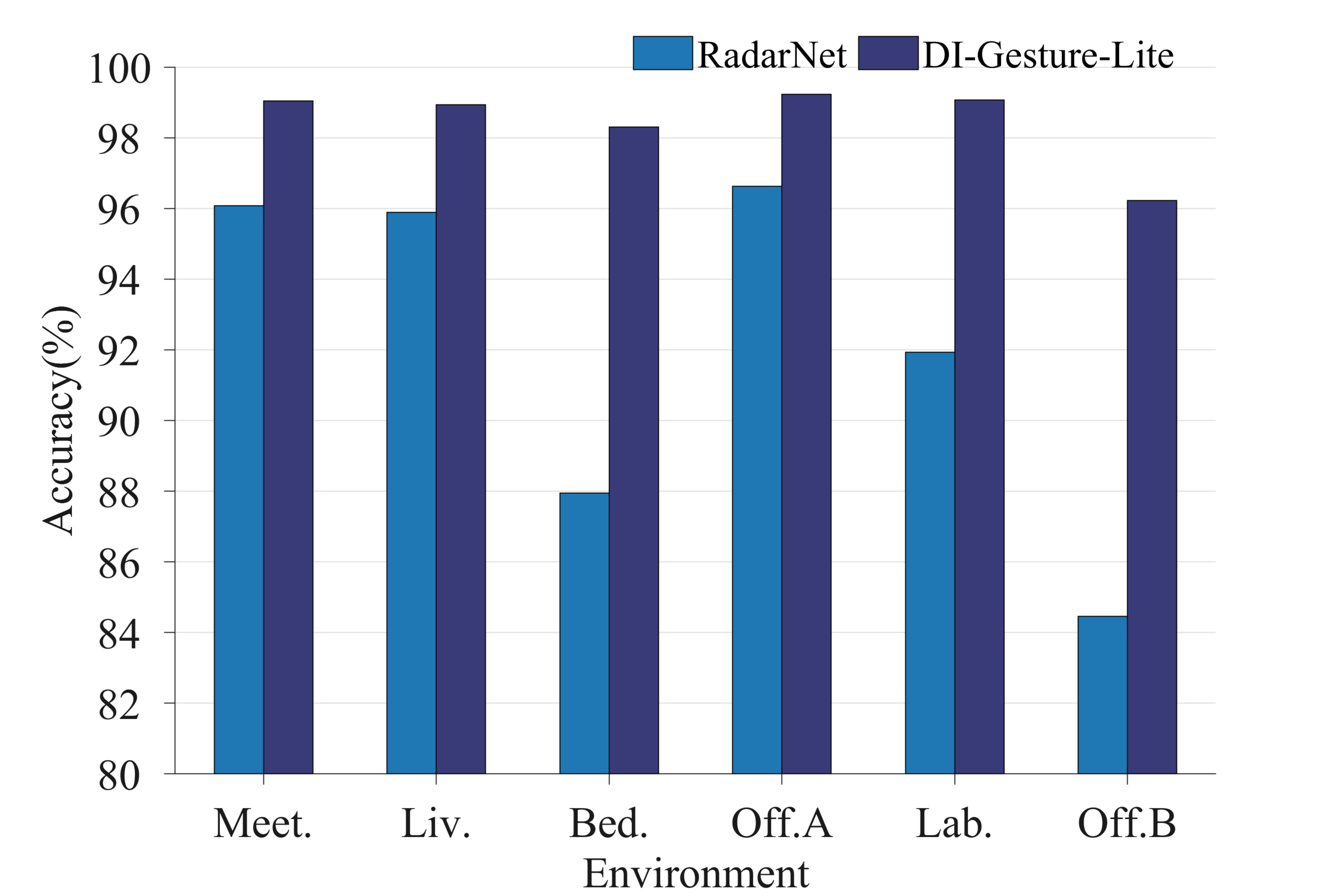}
		\caption{Comparison of new room test.}
		\label{comparenewenv}
	\end{minipage}%
	\begin{minipage}[t]{0.33\textwidth}
		\centering
		\includegraphics[scale= 0.065]{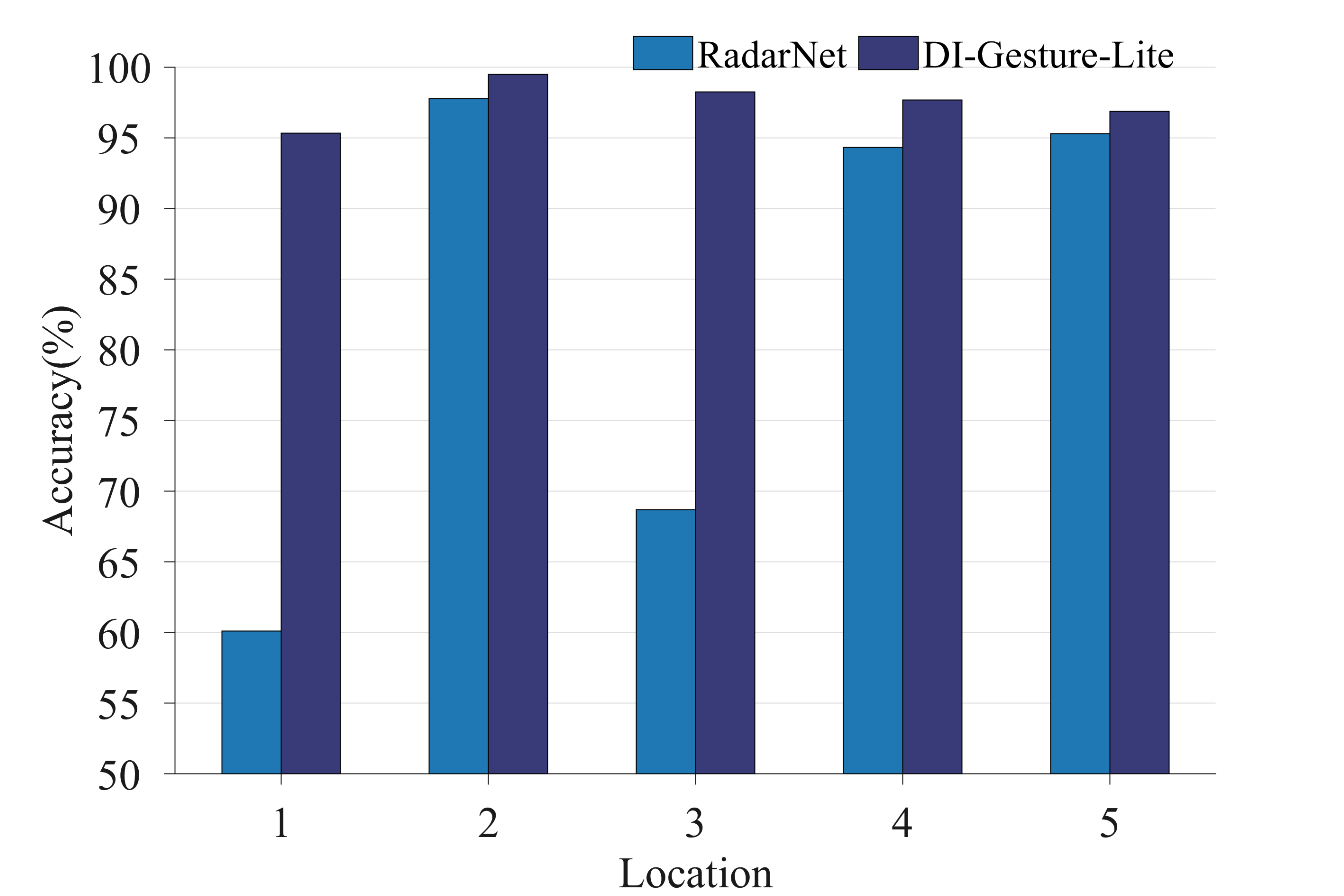}
		\caption{Comparison of new location test.}
		\label{comparenewloc}
	\end{minipage}%
	\quad
	\label{envloc}
\end{figure*}

We first compare the in-domain and cross-domain performance of DI-Gesture-Lite with RadarNet and then make a detailed analysis on why DI-Gesture-Lite performs much better than RadarNet on new domains. In this experiment, the partition of the training set and test set for in-domain, new user, new environment and new location test are same as Section 4.3. As can be observed from Table \ref{comparison}, both two methods work quite well on familiar domains. However, RadarNet's performance on new domains significantly drops to 88.64\% for new users, 92.16\% for new environments, and 83.24\% for new locations. In contrast, DI-Gesture-Lite can still preserve an impressive accuracy of 97.05\%, 98.47\%, and 97.51\% for new users, new environments, and new locations, respectively. To better understand the performance gap between the two solutions, we present the recognition accuracy for each gesture of the new user test in Fig. \ref{comparenewuser}.
Compared with RDI, DRAI filters signals reflected from human body parts(e.g. torso, leg) which are relatively static and other parts(e.g. head, shoulder) with subtle motion compared to hand. Therefore, DI-Gesture-Lite is more robust to new users and can achieve better performance than RadarNet on each gesture. Fig. \ref{comparenewenv} shows the accuracy of two solutions in different environments. It is clear that the performance of DI-Gesture-Lite surpasses RadarNet in each environment because DRAI removes static clutter and reduces the influence of the multipath effect, thus is more robust to environmental changes. As for different locations, RadarNet is able to perform well at location 2, 3, and 5 but cannot generalize to location 1 and 3, as shown in Fig. \ref{comparenewloc}. This is because that location 2, 3, and 5 have different angles relative to the radar while location 1 and 3 are at different ranges away from the radar. Therefore, gestures performed at location 1 and 3 show significant differences due to the fine-grained range resolution of RDI. Instead, DI-Gesture-Lite solves this problem by the data augmentation technique which can synthesize gesture data at different locations thus achieving high accuracy on different locations.

\begin{figure*}[htbp]
	\begin{minipage}[t]{0.3\textwidth}
		\centering
		\includegraphics[scale= 0.065]{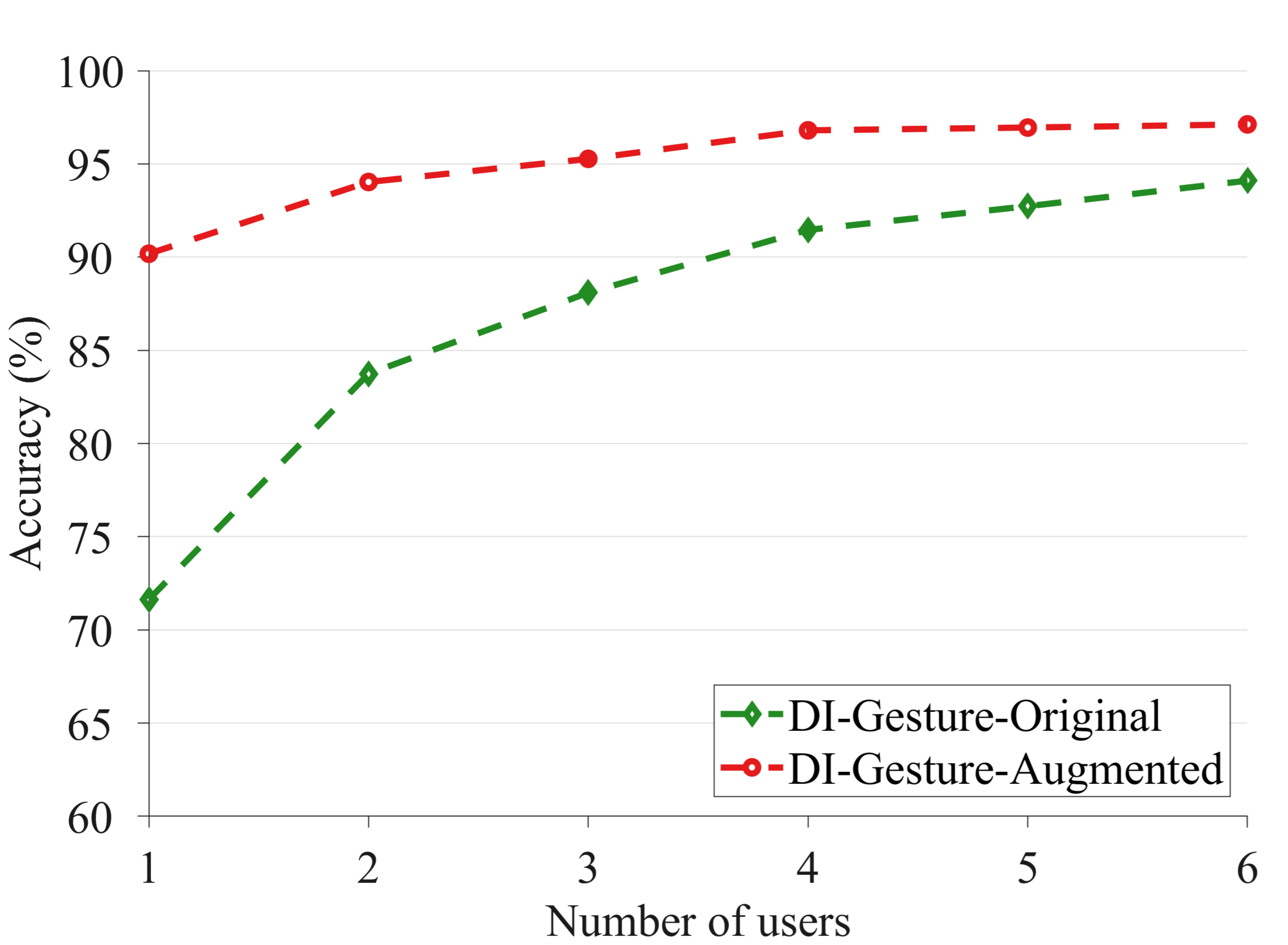}
		\caption{Impact of data augmentation on cross-user test.}
		\label{augmentationuser}
	\end{minipage}%
\qquad
	\begin{minipage}[t]{0.3\textwidth}
		\centering
		\includegraphics[scale= 0.065]{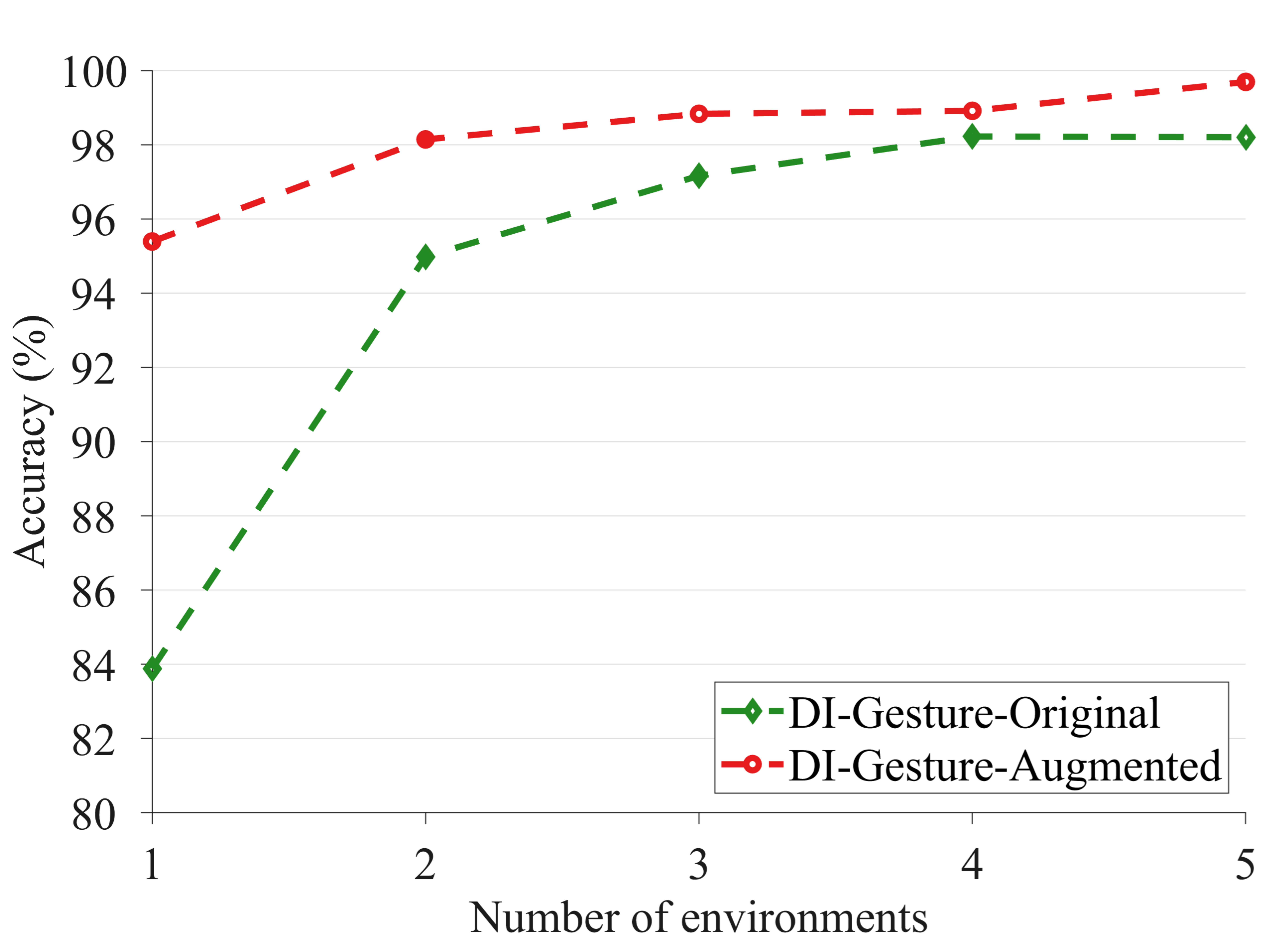}
		\caption{Impact of data augmentation on cross-environment test.}
		\label{augmentationenv}
	\end{minipage}%
\qquad
	\begin{minipage}[t]{0.3\textwidth}
		\centering
		\includegraphics[scale= 0.065]{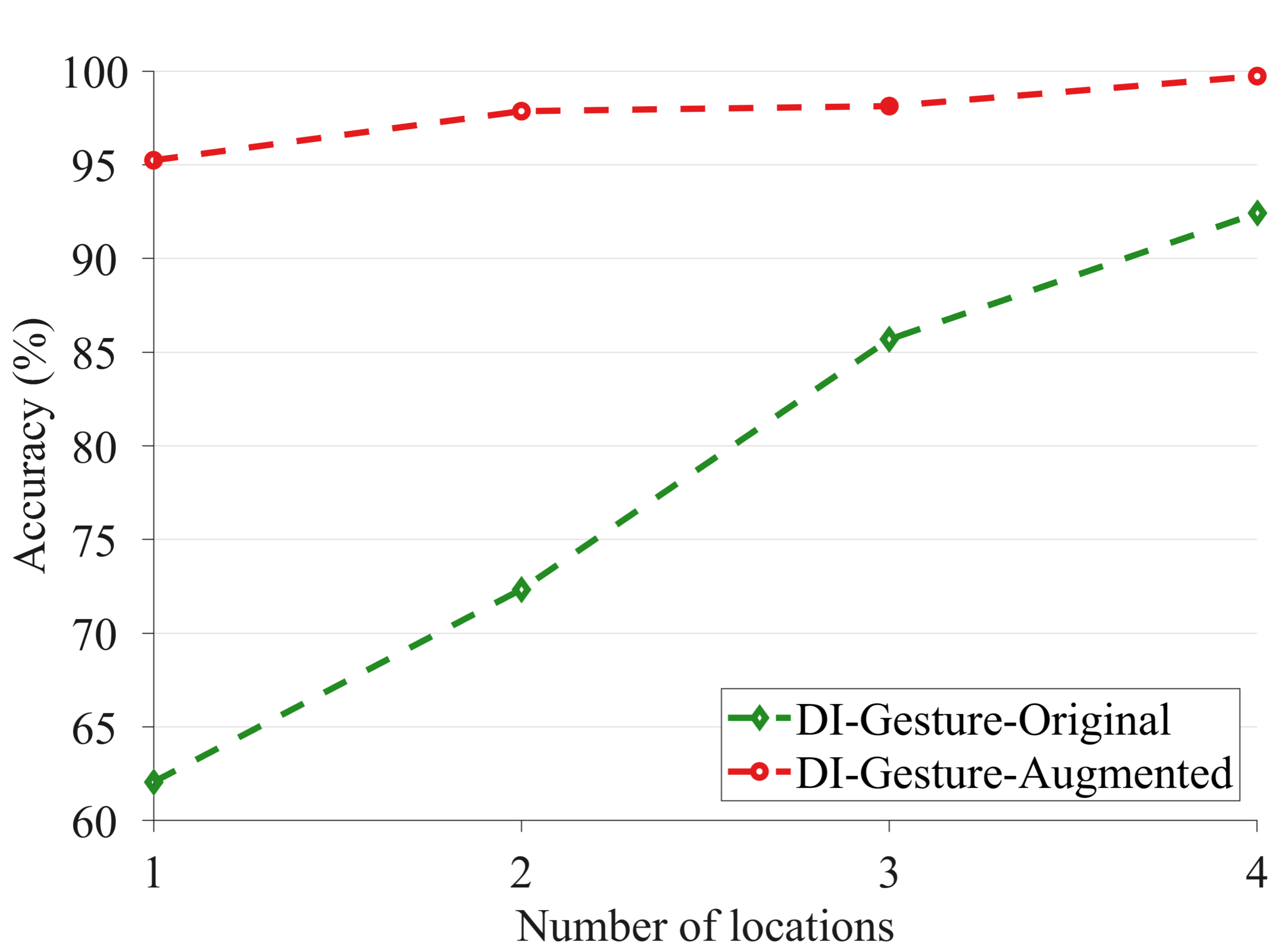}
		\caption{Impact of data augmentation on cross-location test.}
		\label{augmentationloc}
	\end{minipage}%
	\quad
	\label{envloc}
	\vspace{-0.5cm}
\end{figure*}
\subsection{Impact of Data Augmentation}
\subsubsection{Sensing at new domains}
The robustness of the deep learning model is closely related to the number of domains that the model has seen. To evaluate the effect of the proposed data augmentation technique, we train DI-Gesture with and without data augmentation, denoted as DI-Gesture-Augmented and DI-Gesture-Original, respectively. Besides, we gradually reduce the number of domains in the training set to simulate situations where sufficient training data is not available. The impact of data augmentation on new users, new environments, and new locations is shown in Fig. \ref{augmentationuser}, Fig. \ref{augmentationenv} and Fig. \ref{augmentationloc}, respectively. From these results, we have two key observations: $(i)$ DI-Gesture-Augmented outperforms DI-Gesture-Original in all experimental settings, which demonstrates that the proposed data augmentation methods indeed provide more qualified data to reduce the influence of gesture inconsistency in different domains and enhance system robustness.  $(ii)$ The fewer domains appear in the training set, the more improvement of recognition accuracy achieved after training with augmented data. This result indicates that the data augmentation technique can work well when only limited collected data is available, which is the most common situation in practice.  

\begin{figure}[htbp]
	\begin{center}
		\includegraphics[scale=0.09]{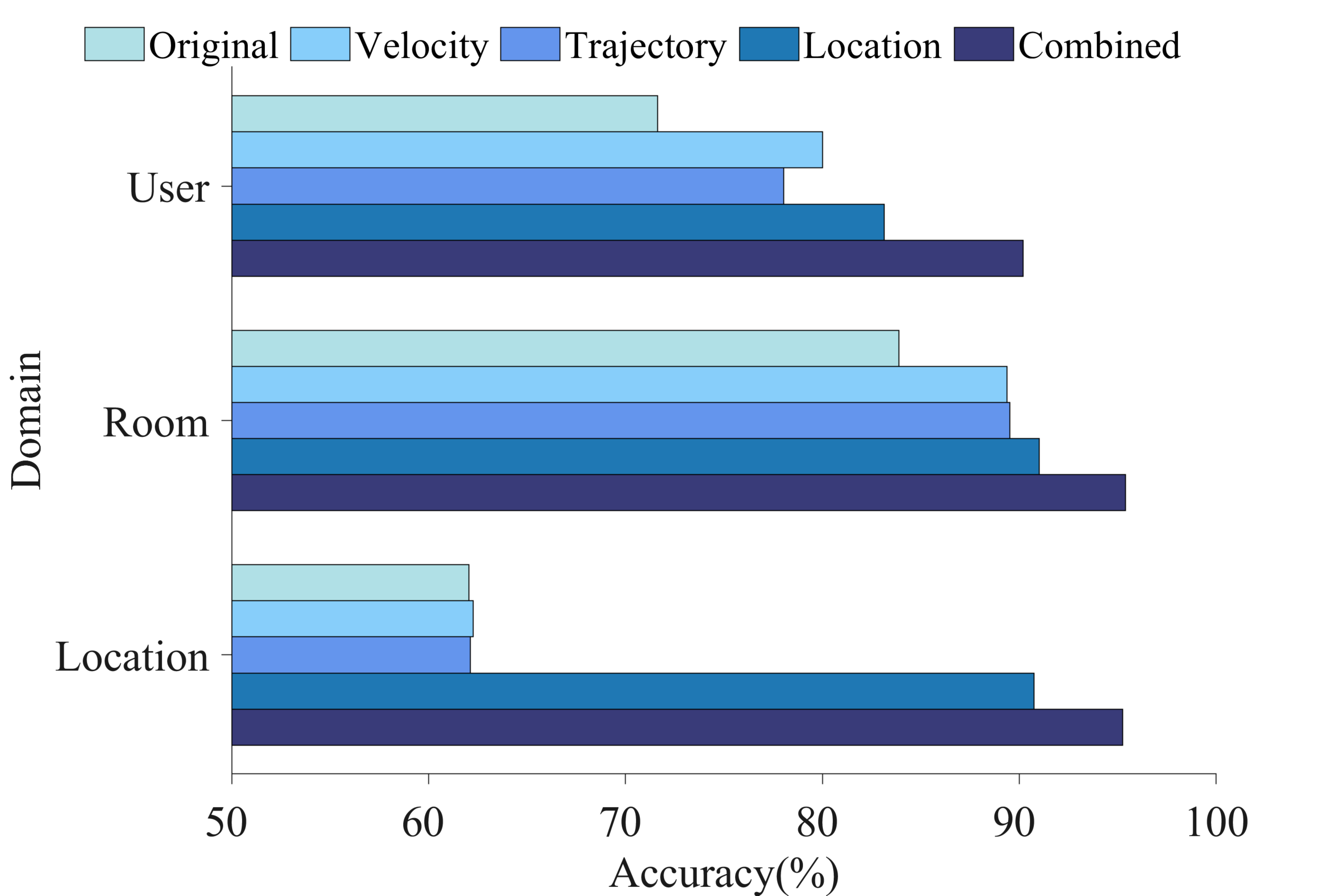}
		\caption{Impact of each data augmentation method in different cross-domain scenarios.}
		\label{eachaug}
	\end{center}
\end{figure}
To validate the effectiveness of each data augmentation method, we combine the data augmentation for different distances and angles as location augmentation and train DI-Gesture with each of the three different augmentation methods. To explore the impact of data augmentation on insufficient samples, we only use data collected from 1 user, 1 environment, and 1 location as the training data and test the rest of the data. The results are shown in Fig. \ref{eachaug}, the "original" denotes the recognition accuracy of DI-Gesture without data augmentation, and the "combined" represents the accuracy of DI-Gesture trained with the combination of the three augmentation methods. From Fig. \ref{eachaug}, we can observe that in cross-user and cross-room tests, each of the augmentation methods can improve the accuracy by at least 5.48\%, and the location augmentation can increase the accuracy of new user test from 71.63\% to 83.14\%. This is because even if standing at the same anchor location, different users usually perform gestures with different scales and directions, resulting in range and angle variations in DRAI. In cross-location test, the location augmentation can even bring a 28.68\% accuracy improvement comparing DI-Gesture-Original without data augmentation, since the location difference is the main reason for pattern variations of DRAI. Another key observation is that when DI-Gesture is trained with the combination of three augmentation methods, the accuracy of all different cross-domain settings can reach over 90\%, which demonstrates the effectiveness of the data augmentation framework.

\subsubsection{Sensing at extreme angles}
To validate the effectiveness of extended data augmentation techniques in extreme-angle scenarios, we collect 600 gesture samples from 10 different users at $\pm$45° and $\pm$60°, respectively. We take gesture samples collected at 0°, 80cm in the released dataset as the training set, which has 2120 samples in total.
We firstly train the model by simply shifting the DRAI along the angular domain as a baseline, then we evaluate the impact of the proposed new data augmentation methods by individually combining them with shifting. As shown in Table \ref{new_aug_result}, each of the data augmentation methods brings a noticeable improvement in recognition accuracy and the improvement is more significant for higher angular displacement.
When training the model with the combination of all three data augmentation methods, the recognition accuracy of gestures performed at 45° and
60° reaches 97.33\% and 93.83\%, which is 7.66\% and 22.50\% higher than the baseline, respectively. 
\begin{table}[htbp]
	\centering
	\caption{Impact of data augmentation in the extreme-angle scenario.}
	\begin{tabular}{ccc}
		\toprule
		Data augmentation & ±45° (\%) & ±60° (\%) \\
		\midrule
		Different angles & 89.67 & 71.33 \\
		Different radiated power & \textbf{91.83 (+2.16)} & \textbf{89.33 (+18.00)} \\
		Different angular resolutions & \textbf{92.50 (+2.83)} & \textbf{82.17 (+10.84)} \\
		Different geometric features & \textbf{92.67 (+3.00)} & \textbf{81.33 (+10.00)} \\
		Combined & \textbf{97.33 (+7.66)} & \textbf{93.83 (+22.50)} \\
		\bottomrule
	\end{tabular}%
	\label{new_aug_result}%
\end{table}%

We have also noted that with the increase of angular displacement, gesture movements are more likely to be out of the radar's field of view and trajectories of different gestures become more ambiguous. This observation probably explains why the accuracy at  $\pm$60° is lower than  $\pm$45°. It also indicates that the performance would be further limited at extremely higher angular displacements (i.e. larger than 60°).

\subsection{Performance in Real-time Scenario}
In this section, we evaluate the performance of DI-Gesture in the real-time scenario, including the recognition ability when users  perform multiple gestures continuously in the single-person scenario, the recognition accuracy in the presence of other interfering users, and the computational consumption when the system runs in real-time.
\subsubsection{Performance in Single-person Scenario}
In unsegmented recognition tasks, we use two metrics for performance evaluation, including continuous recognition accuracy (CRA) and multiple prediction rate (MPR). For CRA, the recognition result is wrong when the gesture is misclassified or there is no prediction, which can be denoted as
\begin{equation}
	CRA = 1-\frac{W+M}{N},
\end{equation}
where $W$ is the number of gestures which are misclassified, $M$ is the number of gestures that the system do not make any prediction and $N$ is the actual number of gestures that the user performed. Besides, the MPR requires the system to make only one prediction for each gesture. Suppose that  $P$ is the number of predictions that the system output,  the MPR can be expressed as

\begin{equation}
	MPR = 1-\frac{N}{P}
\end{equation}

To validate the real-time recognition ability of DI-Gesture, we train DI-Gesture with data collected from 6 environments, 4 locations, and 20 users. Then we ask 8 users (i.e. 4 familiar users and 4 new users) standing at the rest of 1 unseen location in a new environment, which is a more practical and challenging situation. The users continuously perform each predefined gesture 10 times and the system uses the dynamic window mechanism and the classifier to segment and recognize each gesture automatically. We also implement the fixed-length sliding window approach similar to \cite{radarnet} and use the same classifier for a fair comparison. The results are shown in Table \ref{mdr1} and Table \ref{mdr2}. We can observe that the proposed dynamic window mechanism achieves higher CRA and much lower MPR compared with the fixed-length sliding window. We believe this is because that fixed-length sliding windows can not handle situations where people perform gestures at different speeds. To be specific, if the frame length of the fast gesture is shorter than the step size of the sliding window, it is more difficult to detect. Besides, slow gestures are more often recognized as multiple gestures when their duration is larger than the window size, resulting in the increase of MPR. In contrast, the proposed dynamic window resolves this problem by accurately detecting the start and end of the gesture, and adjusting the size of the recognition window dynamically. The result also demonstrates that DI-Gesture can work well when crossing multiple different domains.
\begin{table}[htbp]
	\centering
	\caption{Comparison of continuous recognition accuracy (CRA) and multiple prediction rate (MPR) for the new room and new location test.}
	\label{mdr1}
	\setlength{\tabcolsep}{4mm}{
	\begin{tabular}{ccc}
		\toprule
		\textbf {Method} & \textbf{CRA (\%)}  &\textbf{MPR (\%)} \\
		\midrule
		Fixed-length sliding window & 95    & 34.25 \\
		Dynamic window mechanism & \textbf{97.08} & \textbf{2.83} \\
		\bottomrule
	\end{tabular}%
}
	\label{tab:addlabel}%
\end{table}%

\begin{table}[htbp]
	\centering
	\caption{Comparison of continuous recognition accuracy (CRA) and multiple prediction rate (MPR) for new room, new location, and new user test.}
	\label{mdr2}
	\setlength{\tabcolsep}{4mm}{
	\begin{tabular}{ccc}
		\toprule
		\textbf {Method} & \textbf{CRA (\%)}  &\textbf{MPR (\%)} \\
		\midrule
		Fixed-length sliding window & 84.58 & 27.93 \\
		Dynamic window mechanism & \textbf{92.91} & \textbf{4.38} \\
		\bottomrule
	\end{tabular}%
}
	\label{tab:addlabel}%
\end{table}%

\subsubsection{Performance in the Presence of Interfering Humans}
\begin{figure}[htbp]
	\begin{center}
		\includegraphics[scale=0.07]{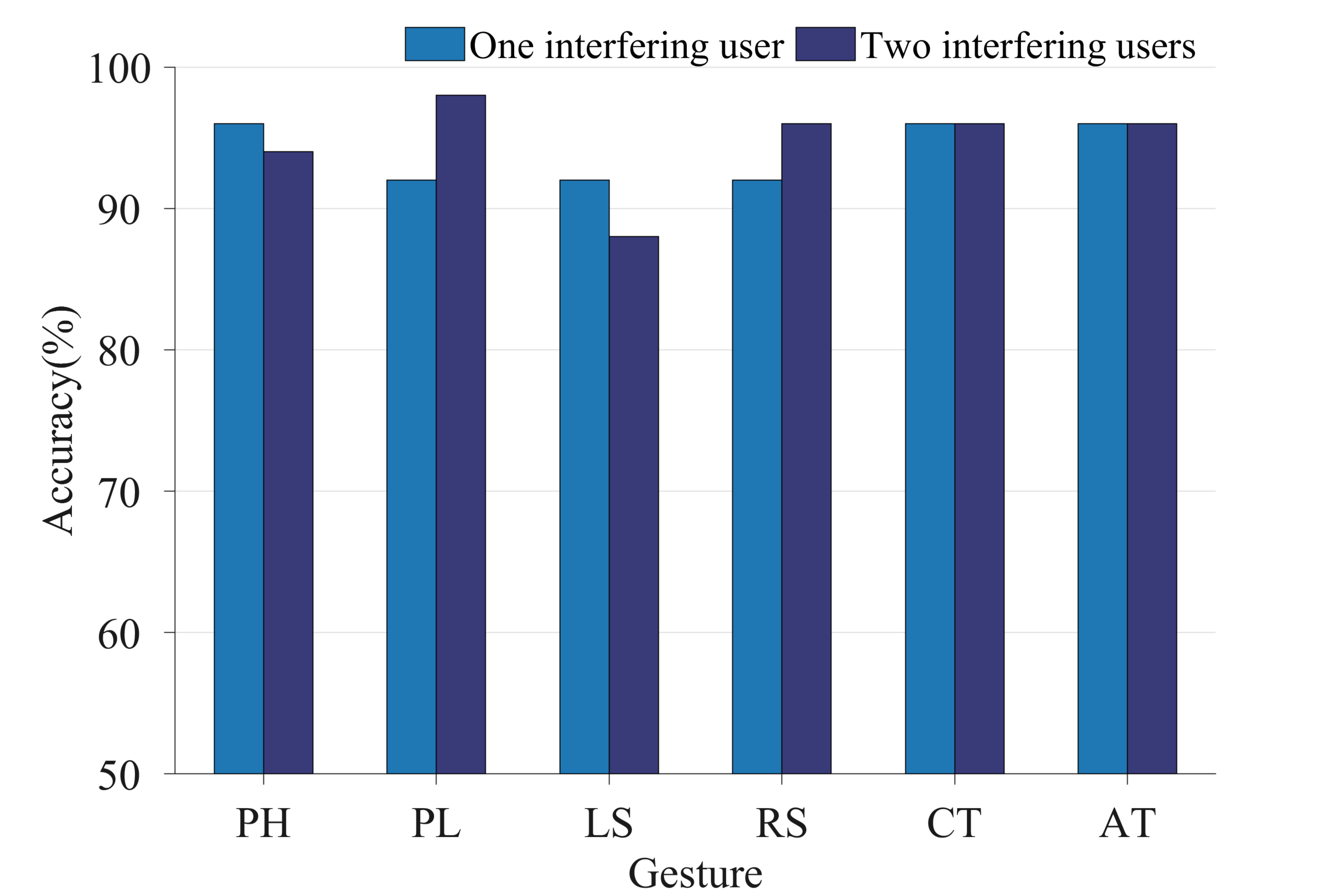}
		\caption{Accuracy of each gesture in the presence of interfering humans.}
		\label{multi_person}
	\end{center}
\end{figure}

To investigate the real-time recognition accuracy in the multiple moving targets scenario, we ask 5 users to perform each gesture for 20 times at different locations. When each user is performing gestures, another one or two interfering persons perform random actions at random locations within the radar detection range (i.e. $\leq$1.5m). As shown in Fig. \ref{multi_person}, the average recognition accuracy in the presence of other interfering users is 94.33\%, which demonstrates the high recognition accuracy of DI-Gesture in multiple moving targets scenario. Since we focus on short-range gesture recognition, the maximum number of interfering users in our experiment is 2 due to limited space. The results also demonstrate that the performance would not be affected by the number of interfering users as the performance with one or two interfering users is similar.

\subsubsection{Computational Consumption}
The computational consumption of DI-Gesture mainly comes from two parts: the signal processing pipeline and the neural network. Hence, to demonstrate that DI-Gesture can run in real-time, we evaluate the signal processing time and the inference time of the neural network, respectively. 

We first run DI-Gesture on a commercial off-the-shelf (COTS) laptop with AMD R7-4800H CPU, and measure the averaged computation time of signal processing steps (i.e. range-doppler-FFT, angle-FFT and nosie elimination) over 2000 frames. As shown in Table. \ref{signal_processs_time}, the computation time of the range-doppler-FFT, angle-FFT, and nosie elimination for each frame is 3.13ms, 1.02ms, and 0.55ms, respectively. Since the frame periodicity of the radar is 50ms, the system is able to finish processing of the current frame before the next frame arrives, thus achieving real-time signal processing.

\begin{table}[htbp]
	\centering
	\caption{Computational consumption of the signal processing steps for each frame.}
	\begin{tabular}{cc}
		\toprule
		\textbf{Signal processing step} & \textbf{Computation time (ms)} \\
		\midrule
		Range-Doppler FFT & 3.13 \\
		Angle FFT & 1.02 \\
		Noise elimination & 0.55 \\
		Total & 4.7 \\
		\bottomrule
	\end{tabular}%
	\label{signal_processs_time}%
\end{table}%

\begin{table}[htbp]
	\centering
	\caption{Comparison of the model size and inference time.}
	\label{time}
	\setlength{\tabcolsep}{4mm}{
		\begin{tabular}{ccc}
			\toprule
			\textbf{Model} & \textbf{Model size (MB)} & \textbf{Inference time (ms)} \\
			\midrule
			DI-Gesture-Lite & 0.16  & 2.45 \\
			DI-Gesture & 0.69  & 2.87 \\
			\bottomrule
		\end{tabular}%
	}
	\label{tab:addlabel}%
\end{table}%
 Next, we evaluate the model size and inference time of DI-Gesture and DI-Gesture-Lite. We implement both models on the same laptop with CPU only and measure the inference time by taking the average inference time over 1000 runs. As shown in Table \ref{time}, the model size of DI-Gesture-Lite is only 0.16MB and increases to 0.69MB for DI-Gesture due to the larger CNN embedding size and more LSTM hidden nodes. The inference time of DI-Gesture-Lite and DI-Gesture is 2.45ms and 2.87ms, respectively. Compared to DI-Gesture-Lite, the raised inference time of DI-Gesture is 0.42ms, which is negligible. Therefore, both DI-Gesture-Lite and DI-Gesture are small and fast enough to be implemented in real-time scenarios.

\section{Limitation}
In this paper, we address the domain dependence problem of mmWave gesture recognition. The range of gestures in our experiment is currently limited to 1m.
Moreover, even if we achieve promising results in the presence of interfering humans, the current spatial segmentation algorithm cannot cope with situations when the interfering user and the actual user are very close (i.e. $<$30cm).  We believe future work in more challenging scenarios like long-range or multi-person gesture recognition will push mmWave sensing further to more applications.

\section{Conclusion}
In this paper, we proposed DI-Gesture, a real-time mmWave gesture recognition system that worked well across new users, new environments, and new locations.
DI-Gesture outperformed the state-of-the-art in two aspects: $(i)$ The proposed data augmentation framework for mmWave signals enabled an impressive cross-domain accuracy without collecting extra data or model retraining;  $(ii)$ The spatial-temporal gesture segmentation helped DI-Gesture achieve a more satisfied performance when the system worked in real-time. Furthermore, we collected the first cross-domain mmWave gesture dataset consisting of 24050 gesture samples from 25 volunteers, 6 environments, and 5 locations and made it public to the research community. We believe that the proposed methods and released dataset not only push the mmWave gesture recognition into real-world applications, but also can be applied to other wireless sensing tasks and inspire more researchers to investigate this ubiquitous sensing technique.

%\appendices
%\section{}
%\ifCLASSOPTIONcompsoc
%\else
%\fi

\ifCLASSOPTIONcaptionsoff
  \newpage
\fi

\small
\bibliographystyle{IEEEtran} %声明选择的格式
\bibliography{refs} %bib文件名，需要放在同一个文件夹下，否则要在filename前说明路径

\end{document}